%% file: paper.tex
\documentclass[twoside]{article}

\newcommand{\ourmaintitle}{Mapping and Advancing the Scalability--Accuracy Frontier of Nonlinear Causal Discovery}

\newcommand{\ourtitle}{\ourmaintitle}
\newcommand{\ourmethod}{\textsc{Spade}\xspace}
\newcommand{\ourmethodGauss}{\textsc{Spade-Gauss}\xspace}
\newcommand{\ourmethodHist}{\textsc{Spade-Hist}\xspace}

\RequirePackage{amsmath}
\RequirePackage{amssymb}
\PassOptionsToPackage{numbers,sort&compress}{natbib}

\input{preamble}

\RequirePackage{mathtools}
\mathtoolsset{showonlyrefs}
\usepackage[preprint]{AISTATS2027PaperPack/aistats2027}
\usepackage[hyphens]{url}
\usepackage{graphicx}
\usepackage{natbib}

\usepackage{caption}
\begin{document}

\twocolumn[

\aistatstitle{\ourtitle}

% \aistatsauthor{Hendrik Suhr \And Sascha Xu \And Jilles Vreeken}

\aistatsauthor{
Hendrik Suhr
\And
Sascha Xu
\And
Jilles Vreeken
}

\aistatsaddress{
 CISPA Helmholtz Center for Information Security\\
\{\texttt{hendrik.suhr},\texttt{sascha.xu},\texttt{jv}\}\texttt{@cispa.de}
}
]

\begin{abstract}
\input{abstract}
\end{abstract}

\input{introduction}

\input{preliminaries}

\input{theory}

\input{instantiation}

\input{experiments}

\input{conclusion}

\clearpage
\input{ai_use_statement}

\bibliography{bib/abbreviations,bib/bib-paper}
\bibliographystyle{abbrvnat}

\clearpage
\input{checklist}
\clearpage
%% apx
\ifapx
\appendix
\onecolumn
\include{appendix}
\fi

\end{document}

%% file: preamble.tex
\newif\ifapx
\apxtrue % uncomment if you want appendix
\newif\ifpaper 
\paperfalse
\usepackage[dvipsnames]{xcolor}
\usepackage{xspace}
\usepackage{diagbox} % put this in your preamble
\usepackage{bigints}
\usepackage{algorithm}
\usepackage{algorithmic}
\usepackage{latexsym}
\usepackage{subcaption}
\usepackage{graphicx,centernot}

\usepackage{footmisc}
\usepackage{pgfplotstable}
\usepackage{pgfplotstable}

\usepackage{dblfloatfix}
\usepackage{afterpage}
\usepackage{pgfplots}
\usepackage{pgfplotstable}
\usepackage{booktabs}
\usetikzlibrary{patterns}
\usepackage{tabularx, booktabs}
\usepackage{dsfont}
\pgfplotsset{compat=1.18}
\usepgfplotslibrary{groupplots}

\graphicspath{ {./figs/} }

\usetikzlibrary{arrows.meta, positioning,shapes.geometric, arrows}

\input{defines}
\usetikzlibrary{pgfplots.fillbetween}
\usepackage{amssymb}  % <- provides \lightning
\usepackage{fontawesome5}

\usetikzlibrary{patterns}

%% file: defines.tex
\usepackage[most]{tcolorbox}
\definecolor{slate}{RGB}{44, 62, 80}

\newtcolorbox{desiderata}[1][]{
    enhanced,
    colback=gray!5,            % Subtle background tint
    colframe=slate,            % Custom left border color (defined below)
    arc=0mm,                   % Sharp corners for an academic look
    boxrule=0pt,               % Remove border around the box
    leftrule=3pt,              % Add a thick left accent border
    coltitle=black,            % Title color
    fonttitle=\bfseries\large, 
    attach title to upper,
    after title={\par\medskip},
    left=4mm, right=4mm,       % Padding inside the box
    top=3mm, bottom=3mm,
    #1                         % Allows passing custom overrides
}

\definecolor{OIblue}{HTML}{0072B2}
\definecolor{OIvermillion}{HTML}{D55E00}

\makeatletter
\newcommand\resetstackedplots{
\makeatletter
\pgfplots@stacked@isfirstplottrue
\makeatother
\addplot [forget plot,draw=none] coordinates{AP};
}
\makeatother

\newcommand{\TODO}[1]{%
  \textcolor{red}{\textbf{TODO:} #1}%
  \xspace
}

\newcommand{\Xb}{\ensuremath{\mathbf{X}}}

\newcommand{\Gc}{{\ensuremath{\mathcal{G}}}}

\DeclareMathOperator{\Pa}{Pa}

\newcommand{\PaGj}{\Pa_j^{\Gc}}
\newcommand{\scoref}{\mathcal{S}}

\newcommand{\argmin}{\operatorname{argmin}}
\newcommand{\CAPS}{\textsc{Caps}\xspace}

\newcommand{\TOPIC}{\textsc{Topic}\xspace}
\newcommand{\CAM}{\textsc{CAM}\xspace}
\newcommand{\NOGAM}{\textsc{NoGAM}\xspace}
\newcommand{\SCORE}{\textsc{Score}\xspace}
\newcommand{\PC}{\textsc{PC}\xspace}
\newcommand{\FCI}{\textsc{FCI}\xspace}
\newcommand{\DIFFAN}{\textsc{DiffAN}\xspace}
\newcommand{\SCINO}{\textsc{SciNO}\xspace}
\newcommand{\AVICI}{\textsc{Avici}\xspace}
\newcommand{\CAUSCALE}{\textsc{CauScale}\xspace}
\newcommand{\DAGMA}{\textsc{Dagma}\xspace}
\newcommand{\SDCD}{\textsc{SDCD}\xspace}
\newcommand{\NOTEARS}{\textsc{NoTears}\xspace}

\newcommand{\DAS}{{\textsc{DAS}}\xspace}

\newcommand{\GLOBE}{\textsc{Globe}\xspace}

\definecolor{OurColor}{RGB}{0,180,0}        % bright green (keep)
\definecolor{OurCAMColor}{RGB}{0,100,0}     % dark green (keep)

\definecolor{AEColor}{RGB}{230,25,75}       % vivid raspberry red
\definecolor{NFColor}{RGB}{245,130,48}      % bright tangerine
\definecolor{LOFColor}{RGB}{255,225,25}     % sunshine yellow
\definecolor{IsoColor}{RGB}{60,180,220}     % sky blue
\definecolor{DeepSVDDColor}{RGB}{145,30,180} % purple
\definecolor{CopodColor}{RGB}{255,190,200}  % soft pink

\definecolor{SOColor}{RGB}{240,50,230}      % magenta
\definecolor{DICEColor}{RGB}{0,130,200}     % teal blue
\definecolor{MOEColor}{RGB}{230,190,255}    % lavender
\definecolor{STColor}{RGB}{245,130,48}     % sunshine yellow
\definecolor{SCMColor}{RGB}{255,225,25}      % hot pink

\definecolor{MarginalColor}{RGB}{245,130,48}     % bright tangerine
\definecolor{MarginalLearnColor}{RGB}{230,25,75} % vivid raspberry red

\definecolor{CAMColor}{RGB}{145,30,180}      % purple
\definecolor{SCOREColor}{RGB}{60,180,220}    % sky blue
\definecolor{NOTEARSColor}{RGB}{255,105,180} % hot pink
\definecolor{NOGAMColor}{RGB}{255,190,200}   % soft pink
\definecolor{TOPICColor}{RGB}{255,225,25}    % sunshine yellow

\DeclareMathOperator*{\argmax}{arg\, max}

%% file: abstract.tex
% Causal discovery from high-dimensional observational data promises to support
% scientific hypothesis generation when interventions are costly or infeasible, but
% nonlinear settings require methods that can extract reliable structural signals while
% searching over large graph spaces. 
Scalable nonlinear causal discovery requires methods that combine flexible mechanism estimators with efficient search over large graph spaces.
Several algorithmic families have been proposed to
address this challenge, yet their accuracy–runtime trade-offs remain poorly
understood. We empirically compare the four major approaches:
differentiable structure learning, amortized structure learning, score-matching, and 
combinatorial search. Our results reveal complementary bottlenecks: differentiable and amortized
methods scale well but exhibit an accuracy gap, score-matching methods
can be accurate in low dimensions but degrade quickly for increasing feature sizes, and
combinatorial methods remain accurate but are slowed by repeated and redundant local 
scoring. 
Motivated by this bottleneck, we develop \ourmethod, a spline-based score-evaluation scheme that compiles  sufficient statistics once and reuses them throughout combinatorial search. Under bounded indegree, its Gaussian variant reduces algorithmic complexity from $O(nd^3)$ to $O(nd^2+d^3)$. Empirically, \ourmethod shifts the observed scalability–accuracy frontier by orders of magnitude: it solves 100-variable problems with 160K samples in seconds and 1600-variable problems with 2.5K samples in minutes, while retaining high structural accuracy across synthetic and real-world benchmarks. These results reveal a substantial shift in the practical scale of combinatorial search and highlight the importance of evaluating scalable causal-discovery methods along the full accuracy–runtime frontier.
% We hope that our findings will guide future research toward scalable and accurate causal discovery in high-dimensional nonlinear settings.
% Overall, \ourmethod advances the of nonlinear causal discovery,
% retaining the structural accuracy of combinatorial search while making repeated
% nonlinear scoring computationally feasible at much larger scales.

%% file: introduction.tex
\section{Introduction}
\begin{figure}[b]
    \centering
    \includegraphics[width=0.9\linewidth]{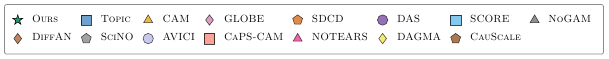}
 \begin{subfigure}{0.49\linewidth}
        \centering
        \includegraphics[width=\linewidth]{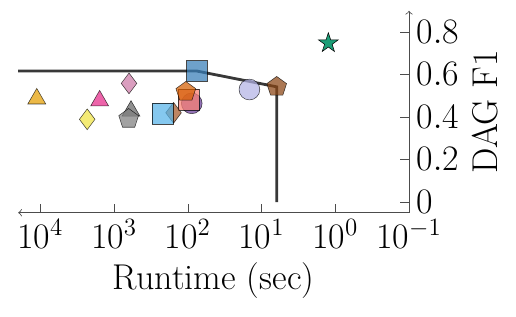}
        \caption{Scalability--accuracy frontier on synthetic data.}
        \label{fig:synthetic_pareto}
    \end{subfigure}%
    \hfill
    \begin{subfigure}{0.49\linewidth}
        \centering
        \includegraphics[width=\linewidth]{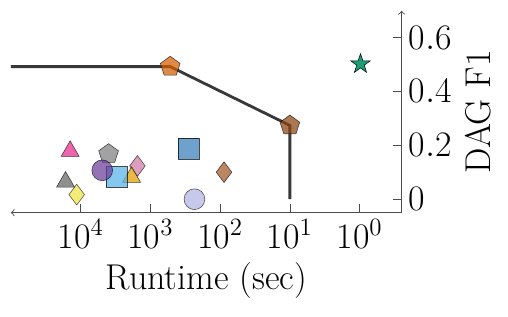}
          \caption{Scalability--accuracy frontier on real-world data.}
          \label{fig:cc_pareto}
    \end{subfigure}%
    % \begin{subfigure}{0.32\textwidth}
    %     \centering
    %     \includegraphics[width=\linewidth]{figs/tikz/sachs_pareto_frontier.pdf}
    % \end{subfigure}%
   
\caption{Scalability--accuracy Pareto comparison on the synthetic qualitative benchmark,
with F1 scores and runtimes averaged over 480 datasets (left), and on the
real-world Causal Chamber benchmark (right).}
\label{fig:pareto}
\end{figure}%
 
\label{sec:intro}
Learning causal structure from observational data is a central problem in science
and machine learning. In many applications, interventions are costly, infeasible,
or unethical, whereas large observational datasets are readily available. Causal
discovery can support scientific hypothesis generation, improve robustness under
distribution shift, and guide downstream decisions with causal
rather than predictive relationships.

%Causal discovery spans a broad range of assumptions, identifiability regimes, and
%algorithmic strategies; we refer to \citet{zanga:2022:survey} for an overview.
In this work, we focus on causally sufficient nonlinear additive-noise models, in
which the directed acyclic graph (DAG) is generically identifiable from
observational data rather than only up to a Markov equivalence class
\cite{hoyer:2008:nonlinear,peters:2014:identifiability}. 
Yet identifiability
does not make high-dimensional discovery easy: the space of DAGs grows
super-exponentially, finite samples make structure selection uncertain, and
nonlinear mechanisms typically require flexible conditional models rather than
simple closed-form scores.

Four main algorithmic families have been proposed to tackle this challenge.
Combinatorial score-based methods assign DAGs or causal orders a likelihood score and search for high-scoring solutions \cite{chickering2002optimal,buhlmann:2014:cam,mian:2021:globe,xu:2025:topic}. 
Score-matching methods exploit differential properties of the observational density, replacing explicit graph search by estimation of the Hessian of the log density \cite{rolland:2022:score,montagna:2023:nogam,sanchez:2023:diffan,kang:2026:scino}. 
Differentiable structure-learning methods relax the discrete adjacency matrix to continuous parameters and impose acyclicity through smooth constraints or penalties \cite{zheng:2018:notears,2020:lachapelle:grandag,yu:2019:dag,nazaret:2024:stable}. 
Amortized causal discovery uses transformers that are pretrained on synthetic causal models, which act as a prior, enabling causal discovery with a single forward pass \cite{loewe:2022:acd,lorch:2022:avici,ke:2023:learning,peng:2026:causcale}.
%For a broader overview of causal-discovery methods, we refer to \citet{zanga:2022:survey}.

Despite this progress, it remains unclear how these approaches compare along the
scalability--accuracy frontier, and which bottlenecks prevent accurate methods
from scaling further. Our contributions are threefold:

First, we empirically map the scalability–accuracy frontier of four major families of nonlinear causal-discovery methods as the number of variables and samples increases.

Second, we identify repeated nonlinear local-score evaluation as a key bottleneck limiting the scalability of otherwise accurate combinatorial methods, motivating a reusable-score approach to graph search.

Third, we instantiate this approach in \ourmethod, which combines a spline-accelerated local score with the \TOPIC search algorithm~\citep{xu:2025:topic}. By compiling reusable sufficient statistics for additive-regression scores, \ourmethod reduces repeated nonlinear scoring during sparse order search and substantially improves the observed scalability–accuracy frontier (Fig. \ref{fig:pareto}).

%% file: preliminaries.tex
\section{Preliminaries}
\label{sec:prelim}

We observe an i.i.d.\ sample
\(\Xb=(x^{(1)},\ldots,x^{(n)})^\top \in \mathbb{R}^{n\times d}\)
from a distribution \(P_X\) over random variables \(X_1,\ldots,X_d\). We assume
that \(P_X\) is generated by a causally sufficient structural causal model
(SCM) with directed acyclic graph (DAG) \(\Gc=(V,E)\), where
\(V=\{1,\ldots,d\}\). An edge \(i\to j\in E\) indicates that \(X_i\) is a direct
cause of \(X_j\). We denote the causal parents of \(X_j\) by
$
  \Pa_j := \{ i\in V : i\to j\in E \},
$
and write \(X_{\Pa_j}:=(X_i)_{i\in\Pa_j}\). A DAG induces the usual recursive
factorization
\begin{equation}
  p(x)
  =
  \prod_{j=1}^d p\!\left(x_j \mid x_{\Pa_j}\right).
  \label{eq:dag-factorization}
\end{equation}
Thus, score-based causal discovery can be viewed as searching over acyclic
parent sets \((S_1,\ldots,S_d)\), where \(S_j\subseteq V\setminus\{j\}\) and the
directed graph with edges \(i\to j\) for \(i\in S_j\) is acyclic.

A \emph{causal order} is a permutation \(\pi\) of \(V\) such that all edges are
oriented from earlier to later variables: if \(i\to j\), then
\(\pi^{-1}(i)<\pi^{-1}(j)\). %For a candidate order \(\pi\), the admissible
% parents of \(X_j\) are restricted to its predecessors,
% \[
%   \operatorname{Pred}_\pi(j)
%   :=
%   \{ i\in V : \pi^{-1}(i)<\pi^{-1}(j) \}.
% \]
Orders are therefore a convenient parameterization of acyclic search spaces:
conditioning on \(\pi\) reduces DAG search to selecting, for each node \(j\), a
parent set from its predecessors.

\paragraph{Additive-noise models.}
We focus on additive-noise models (ANMs), in which each variable is generated as
\begin{equation}
  X_j = f_j(X_{\Pa_j}) + N_j,
  \qquad j=1,\ldots,d,
  \label{eq:anm}
\end{equation}
where \(f_j\) is an unknown structural function and
\(N_1,\ldots,N_d\) are jointly independent exogenous noise variables. Under
standard regularity conditions, nonlinear ANMs are generically identifiable
from the observational distribution: except for known exceptional cases, most
notably the linear Gaussian case, an incorrect causal direction does not admit
an ANM representation with independent noise
\citep{hoyer:2008:nonlinear}. Hence,
under these assumptions, the DAG is identifiable from \(P_X\).

In linear Gaussian SEMs, identifiability can be recovered under the
equal-error-variance assumption, which requires all noise variables to have the
same variance \citep{peters:2014:identifiability}. This assumption is fairly restrictive
for heterogeneous real-world data, where quantities are typically recorded on
human-made, non-comparable scales. Thus, we focus only on scale-invariant criteria.

More general identifiable functional causal models have also been studied
\citep{zhang:2009:identifiability,immer:2023:identifiability}. They broaden the
class of identifiable data-generating processes, but also introduce additional
modeling and estimation challenges. We focus on ANMs because scalability already
remains difficult in this simpler fully identifiable nonlinear setting.
For a broader overview of causal-discovery methods, we refer to \citet{zanga:2022:survey}.

%% file: theory.tex
\section{Scalability--Accuracy Trade-Offs}
\label{sec:theory}

We first examine the main algorithmic families proposed for scalable nonlinear causal discovery and inspect their trade-offs in terms of structural accuracy and runtime. 
The goal is to identify the bottleneck that currently limits this trade-off, and thereby to determine which aspect can and should be accelerated.
To this end, we focus on four families of nonlinear causal-discovery methods: combinatorial score optimization, differentiable structure learning, amortized structure learning, and score matching. We briefly describe each family here, and provide the exact functioning of each tested method in Appx.~\ref{apx:methods}. 

\paragraph{Combinatorial score optimization.}
Combinatorial methods use discrete optimization to maximize a decomposable score $\scoref$ over candidate DAGs. 
For a graph $\Gc$, with parent set $\PaGj$ for node $j$, a score function commonly takes the form
\begin{equation}
\scoref(\Gc;X)
=
\sum_{j=1}^d s_j(X_j \mid X_{\PaGj})\;,
\label{eq}
\end{equation}
where each local term evaluates how well the proposed parents explain the corresponding child.
In nonlinear additive-noise models, this typically amounts to fitting a local model
\begin{equation}
X_j = f_j(X_{\PaGj}) + N_j
\end{equation}
and scoring the resulting residuals, optionally including a complexity penalty \citep{marx2019identifiability}.

Under suitable model specification, such scores can identify the true DAG in the population limit \citep{marx2022formally}. 
%Their challenge is computational: although local-search heuristics avoid exhaustive enumeration, the algorithms involve many local score evaluations for edge additions, deletions, and pruning.
%When each evaluation involves fitting a nonlinear regression model, these repeated local fits become the computational bottleneck.
We evaluate the order-based methods \CAM~\citep{buhlmann:2014:cam} and \TOPIC~\citep{xu:2025:topic}, as well as the greedy DAG-search method \GLOBE~\citep{mian:2021:globe}.
These methods share related local score structures while exploring the graph space differently.

\paragraph{Differentiable structure learning.}
Differentiable methods replace discrete graph search by continuous optimization over a relaxed adjacency matrix while jointly fitting causal mechanisms.
Here, the objective that is optimized is typically of the form
\begin{equation}
\min_{W,\theta} \; \mathcal{L}(X;W,\theta)
\qquad
\text{s.t.} \qquad h(W)=0\;,
\end{equation}
where $W \in \mathbb{R}^{d \times d}$ is a continuous adjacency matrix, $\theta$ parameterizes the nonlinear mechanisms, and $h(W)=0$ enforces acyclicity.
This formulation turns structure learning into a gradient-based optimization problem and is therefore naturally compatible with flexible neural estimators as well as fast runtime through highly parallelized GPU implementation.

%Its practical challenge is that graph structure and causal mechanisms must be optimized jointly.
%The graph parameters shape the fitted mechanisms, while the fitted mechanisms determine the structural gradients, creating a highly nonconvex objective from which the causal graph has to emerge.
We evaluate nonlinear variants of \NOTEARS~\citep{zheng:2020:learning}, \DAGMA~\citep{bello:2022:dagma}, and \SDCD~\citep{nazaret:2024:stable}, using non-equal-variance objectives where required to match our benchmark setting.
The resulting methods differ in both their mechanism parameterizations and acyclicity formulations.

\paragraph{Score matching.}
Score-matching methods avoid explicit graph search by estimating derivatives of the observational log-density.
Writing
\begin{equation}
\psi(x) = \nabla_x \log p(x)
\end{equation}
for the score function of the data distribution, these methods use score-derived quantities to recover a causal order under suitable assumptions~\citep{rolland:2022:score}.
This replaces optimization over DAGs by density-derivative estimation, and can provide identifiability under comparatively weak assumptions on the functional mechanisms. Some variants also allow more general noise distributions~\citep{montagna:2023:nogam}.
We evaluate \DAS~\citep{montagna:2023:das},
\DIFFAN~\citep{sanchez:2023:diffan}, and
\SCINO~\citep{kang:2026:scino}, focusing on recent score-matching
variants that are explicitly designed for scalable nonlinear causal discovery.
Since \DIFFAN and \SCINO return an order rather than an edge set, we apply
standard \CAM pruning to obtain a DAG estimate.

\paragraph{Amortized causal discovery.}
Amortized causal-discovery methods replace dataset-specific graph search with a learned predictor that maps an entire dataset directly to a distribution over causal graphs.
They are pretrained on simulated SCMs:
\begin{equation}
\phi^\star
=
\arg\min_{\phi}
\;
\mathbb{E}_{(\mathcal{G},X)\sim P_{\mathrm{sim}}}
\left[
\mathcal{L}_{\mathrm{train}}
\bigl(f_{\phi}(X),\mathcal{G}\bigr) 
\right].
\end{equation} 
After pretraining, they enable inference with a single forward pass $\widehat{\mathcal{G}}
=
f_{\phi^\star}(X)$ at test time, shifting computation from per-dataset optimization to offline training~\citep{loewe:2022:acd,lorch:2022:avici,ke:2023:learning,peng:2026:causcale}.

Their inductive biases are therefore determined in part by the simulator distribution, including assumptions on graph structure, mechanisms, noise, and interventions.
We evaluate \AVICI~\citep{lorch:2022:avici} and \CAUSCALE~\citep{peng:2026:causcale}, which provide public pretrained checkpoints and explicitly report generalization to graph sizes beyond their training domain.
% \AVICI performs amortized variational inference over graph structures, whereas \CAUSCALE uses compressed representations and tied attention weights to improve memory and runtime scaling at high dimensionality.

\begin{figure*}[t]
    \centering 
\includegraphics[width=\linewidth]{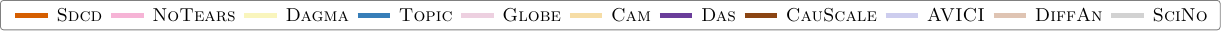}\par
\begin{subfigure}{0.5\linewidth}
\includegraphics[width=0.5\linewidth]{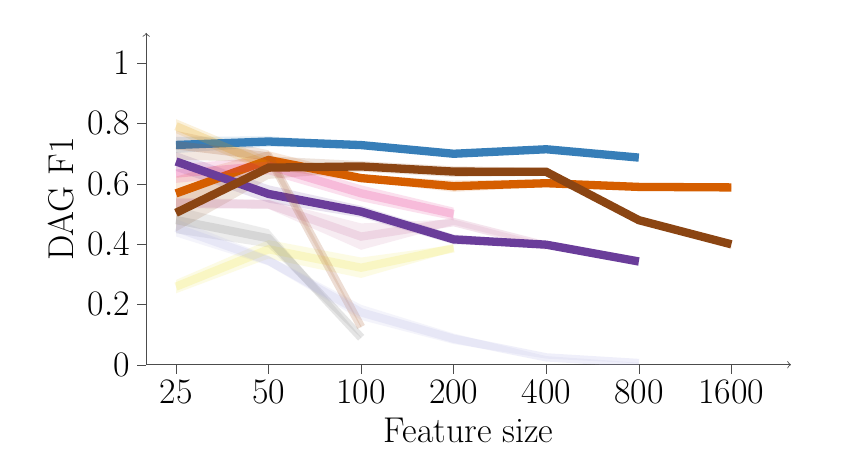}%
\includegraphics[width=0.5\linewidth]{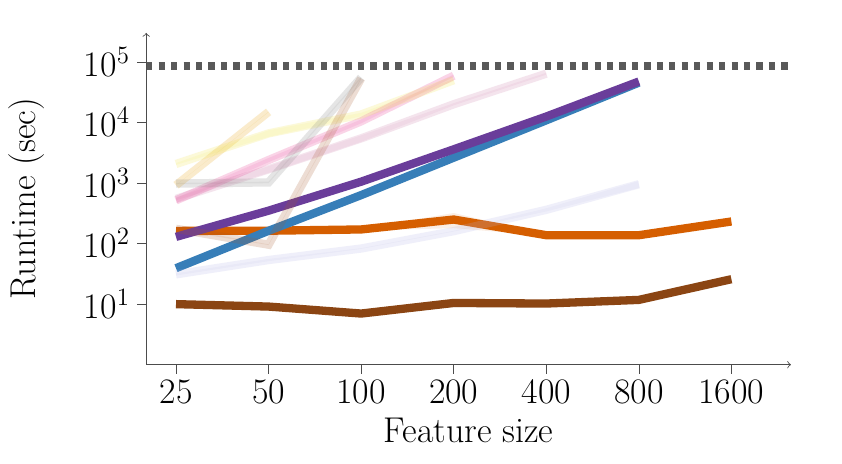}
\caption{Feature complexity.}
\label{fig:benchmark-feature-complexity}
\end{subfigure}%
\begin{subfigure}{0.5\linewidth}
\includegraphics[width=0.5\linewidth]{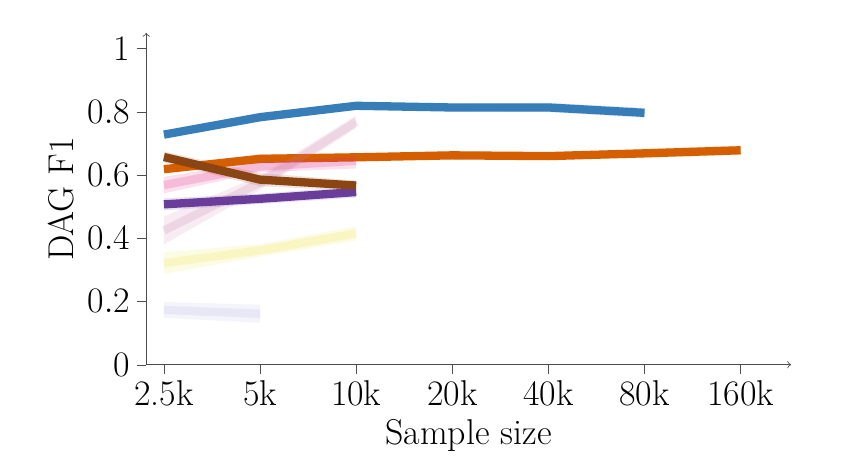}%
\includegraphics[width=0.5\linewidth]{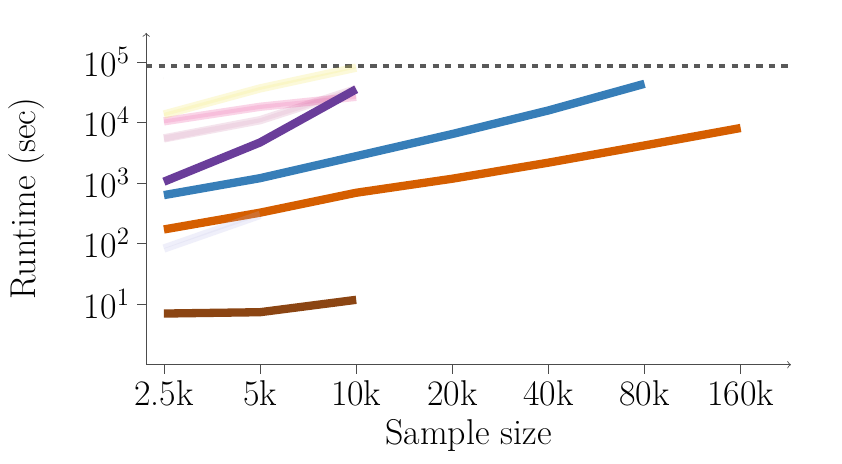}
\caption{Sample complexity.}
\label{fig:benchmark-sample-complexity}
\end{subfigure}%
\caption{Scalability--accuracy trade-offs. Runtime and F1 score are shown in separate
panels with matched dimensions. Dotted lines indicate the 24h timeout. Results are averaged across 5 seeds, shaded regions correspond to standard errors.}
\label{fig:score-matching-samples}
\end{figure*}

\subsection{Experimental Setup}
Our goal is to evaluate how structural accuracy and runtime change as the number of variables and samples increases.
To isolate this scalability--accuracy trade-off, we use a controlled nonlinear additive-noise setting that matches the assumptions of the compared method families as closely as possible.

For the feature-scaling experiment, we vary the number of variables as $d = 25 \cdot 2^k$ and fix the sample size to $n=2{\,}500$.
For each dimension, we generate five random Erd\H{o}s--R'enyi (ER) DAGs with expected degree three and sample one dataset from each DAG.
We increase $k$ until all methods, or all but one method, exceed the time limit of $24$ hours. GPU runs that exceed the 40\,GB device-memory capacity of the available accelerators are recorded as OOM.
For the sample-scaling experiment, we fix $d=100$ and increase the sample size as $n = 2{\,}500 \cdot 2^k$.

We generate nonlinear mechanisms by sampling Gaussian-process functions and use Gaussian exogenous noise.
To avoid exploiting equal-variance or scale cues, we standardize each variable before applying the methods~\citep{ormaniec:2025:standardizing}.
We provide further details of the data-generating process in Appx.~\ref{apx:data}, hardware in Appx.~\ref{apx:hardware}, and implementation details for all methods in Appx.~\ref{apx:implementation_details}.

\subsection{Results}
We present the results in Fig.~\ref{fig:score-matching-samples}. 
For each family, we highlight the representative that achieved the most completed runs: \SDCD for differentiable structure learning, \DAS for score matching, \CAUSCALE for amortized inference, and \TOPIC for combinatorial search. Full metric tables, including precision, recall, SHD, topological divergence, and runtime, are reported in Appx.~\ref{apx:metrics}.

\paragraph{Feature complexity.}
Fig.~\ref{fig:benchmark-feature-complexity} shows a clear separation between runtime scalability and structural accuracy as the number of variables increases.
% Among all methods, the differentiable \SDCD and the amortized \CAUSCALE are able to scale to highest-dimensional problems. For \CAUSCALE, this advantage is due to the single-forward-pass inference of foundation models, whereas the favorable runtime of \SDCD is partly attributable to its quadratic
% acyclicity constraint, in contrast to the cubic matrix operations used by
% \NOTEARS and \DAGMA, and partly to its use of early stopping.
\SDCD and \CAUSCALE scale to the largest feature
dimensions, reflecting the efficient continuous optimization of \SDCD and the
single-forward-pass inference of \CAUSCALE. However, this computational
advantage comes at a substantial cost in structural accuracy: both methods
remain well below the best F1 scores achieved in the benchmark, especially at
the largest feature sizes.
% % The main differentiating factor is that its acyclicity constraint has only quadratic complexity, as compared to the cubic complexity of \NOTEARS and \DAGMA,
% However, this computational advantage comes at a substantial cost in structural accuracy.  \SDCD remains substantially below the highest F1 scores achieved in the benchmark. \CAUSCALE Performs slightly better on the intermediate but significantly worse for the largest feature sizes. 

The highest $F_1$ scores are achieved by the combinatorial method \TOPIC.
Its accuracy degrades only mildly as $d$ grows, indicating that its nonlinear local score retains a strong finite-sample signal even in larger graphs.
The limitation is computational: \TOPIC requires a cubic number of local score evaluations in the number of variables, and each evaluation involves fitting a nonlinear model.
This makes the method substantially more accurate than the differentiable alternatives, but increasingly expensive as the graph grows.

Among the score-matching methods, \DAS gives the strongest feature-scaling performance, due to its efficient pruning.
It is competitive at $d=25$, but its structural accuracy deteriorates rapidly when the number of variables increases.
This suggests that the bottleneck is the estimation of high-dimensional score-derived quantities, which becomes increasingly difficult as the number of variables grows.
The remaining methods either achieve lower accuracy or become substantially slower, timing out between $50$ and $400$ variables.

\paragraph{Sample complexity.}
The sample-scaling experiment in Fig.~\ref{fig:benchmark-sample-complexity} sheds further light on the trade-off between sample size and structural accuracy.
\SDCD remains computationally efficient, but only partially closes the accuracy gap to \TOPIC. 
This suggests that its main limitation might not be sample size alone, but the difficulty of extracting a reliable structural signal from the coupled graph--mechanism objective. We investigate this hypothesis further in Appx.~\ref{apx:experiments}, and the results support this interpretation: Across the considered differentiable methods and regimes, the learned
mechanisms do not reliably rank the true DAG above randomly sampled alternatives.

\CAUSCALE worsens with increasing sample sizes and runs OOM already at 20K samples. This points to a potential mismatch between the pretraining distribution and the benchmark regime, which is a known limitation of amortized approaches \cite{montagna2025demystifying}. Nonetheless, \CAUSCALE remains the most efficient method in its applicable regime.

\DAS improves as the sample size increases, consistent with the benefit of more
accurate estimation of the score-derived quantities used by the method.
However, this improvement comes with substantial runtime cost: the employed
Stein estimator has cubic runtime in the sample size, and \DAS reaches the time
limit at $10$k samples. Thus, \DAS can benefit from additional data but does
not attain the same scalability--accuracy trade-off as combinatorial
search in the present regime.

\TOPIC remains the strongest structural performer as $n$ increases.
This supports the conclusion that nonlinear local scores provide a reliable causal signal when evaluated accurately.
At the same time, larger samples increase the cost of each nonlinear local fit, reinforcing that the bottleneck is not the search signal itself but the repeated evaluation of closely related local models.

\paragraph{Implications.}
The benchmark reveals complementary bottlenecks.
Differentiable and amortized methods offer the best runtime scaling, but lose too much structural accuracy.
Score-matching methods avoid explicit graph search, but their performance is limited by high-dimensional derivative estimation.
Combinatorial score-based search provides the strongest structural signal, but pays for it through repeated nonlinear local model fits.

The latter bottleneck presents a concrete opportunity to improve scalability without sacrificing structural accuracy.
The limitation of \TOPIC is not that the local score lacks signal, but that evaluating it repeatedly is too expensive.
In the next section, we show that for additive spline scores, much of the sample-dependent computation can be cached and reused across parent-set evaluations, enabling significantly faster search.

%% file: instantiation.tex
\section{Scalable Combinatorial Causal Discovery}
\label{sec:algo}

Motivated by the benchmark in the previous section, we target the repeated
local-score evaluations that limit accurate combinatorial methods. The key idea
is to use a nonlinear additive score whose sample-dependent algebra can be
compiled once and reused across the score queries issued during order
search. We instantiate this idea with fixed-basis splines and combine the
resulting score with the order-search framework of \TOPIC. This yields
\ourmethod, an efficient scoring framework that retains likelihood-based nonlinear
scoring while making repeated local-score evaluations computationally feasible.

% This principle is familiar from linear-Gaussian score-based search. In methods
% such as \GES and its scalable variants, local regression scores can be evaluated
% efficiently from covariance or Gram submatrices, which makes repeated local
% score comparisons inexpensive. Nonlinear ANM scores are also local, but they
% usually lose this cheap sufficient-statistic structure because each candidate
% parent set requires fitting a nonlinear mechanism. However, for additive
% splines, analogous structure for nonlinear local scores exists.

% are flexible enough to capture nonlinear additive mechanisms,
% but structured enough that much of the sample-dependent computation can be
% performed once and reused across many candidate parent sets. Importantly, they can be solved with a single closed-form linear regression, if one uses a fixed smoothing parameter instead of optimizing it with iterative fits.  After constructing
% spline bases for each variable, we cache the corresponding Gram blocks,
% feature--response cross-products, and response norms. Scoring a new parent set
% then reduces to assembling and solving a small linear system. For Gaussian
% residual likelihoods, these cached quantities also suffice to evaluate the
% entire local likelihood without materializing residuals.

\subsection{Spline-Accelerated Score Computation}
\label{subsec:score}

We consider a decomposable BIC-type score for candidate parent sets
\citep{schwarz:1978:bic}. During structure search, $\Pa_j$ is unknown, so the
combinatorial search queries local scores for candidate pairs $(j,S)$, with
$S \subseteq V \setminus \{j\}$. For each queried pair, the score fits the
conditional mean of $Y_j$ given $X_S$ and evaluates the resulting residuals by
a penalized negative log-likelihood. For  a residual-likelihood
choice $\ell$, the local score is
\begin{equation}
\mathcal{S}_j(Y_j \mid S;\ell)
=
\mathcal{L}^{\ell}_j(Y_j \mid S)
+
\operatorname{pen}_{\mathrm{BIC}}(S),
\label{eq:local-score}
\end{equation}
where $\mathcal{L}^{\ell}_j(Y_j \mid S)$ is the residual negative
log-likelihood and $\operatorname{pen}_{\mathrm{BIC}}(S)$ is the BIC complexity
penalty.

Decomposable BIC-type scores are a standard choice in score-based causal
discovery. Our contribution is an efficient
scheme to evaluate related local scores repeatedly during combinatorial search, thereby
reducing the overall computational complexity.

\paragraph{Fixed-basis splines.}
We now describe the spline model that exposes the reusable algebra.
We instantiate the conditional-mean model with a fixed-basis generalized
additive spline regression \citep{hastie:1986:gam,wood:2017:gam}.
The basis and smoothing penalty are fixed across score queries, so each queried parent set is scored by
a single penalized least-squares solve rather than by an iterative GAM fit.
This preserves a flexible nonlinear additive model while giving all local regressions a shared feature representation.
% Unlike general-purpose GAM fitting routines, which often solve penalized spline
% models through iterative fitting algorithms, our local mean fit has a fixed
% basis and fixed smoothing penalty. Each candidate parent set is therefore scored
% by solving a single penalized least-squares problem. The fixed basis provides a
% flexible nonlinear additive model while ensuring that all local regressions share
% a common feature representation, which allows their sample-dependent
% computations to be reused across score queries.

For each variable $X_k$, we construct a centered spline design matrix
$B_k \in \mathbb{R}^{n \times K}$, where $K$ is fixed. Let
$Y_j \in \mathbb{R}^n$ denote the centered response for node $j$. For a
candidate parent set $S$, define the concatenated design matrix
$B_S = [B_k]_{k \in S}$. We represent the conditional mean on the observed
sample as
$
f_{j,S}(X_S)
=
B_S \beta_{j,S}.
$
The spline coefficients are estimated by penalized least squares,
\begin{equation}
\widehat{\beta}_{j,S}
=
\arg\min_{\beta}
\left\{
\left\|Y_j - B_S \beta\right\|_2^2
+ 
\beta^\top P_S \beta
\right\},
\label{eq:penalized-spline-fit}
\end{equation}
where $P_S = \lambda \Omega_S + \varepsilon I$.
$\Omega_S$ is the spline roughness-penalty matrix,
$\lambda \geq 0$ controls the degree of smoothing, and
$\varepsilon > 0$ is a small ridge parameter used for numerical stability.
Hence, we obtain the fitted residuals as per
\begin{equation}
\widehat{N}_{j,S}
=
Y_j - B_S \widehat{\beta}_{j,S}\;.
\end{equation}
% We consider two residual likelihoods. The Gaussian variant uses a normal
% residual model with variance estimated from the residual sum of squares. The
% histogram variant estimates a flexible univariate residual density from
% $\widehat{N}_{j,S}$. Both variants use the same additive spline mean model and
% the same BIC penalty; they differ only in how they evaluate the residuals.

\paragraph{Compiling repeated local queries.}
Using combinatorial search, the direct, repeated evaluation of Equation~\eqref{eq:penalized-spline-fit} is the main computational bottleneck.
The same spline bases, response variables, and
overlapping parent sets recur across many local-score queries, yet a naive
implementation repeatedly forms local Gram matrices and feature--response
cross-products from the observations. Because the spline basis and penalty rule
are fixed, this sample-dependent algebra can instead be computed once and
reused.

Let $B = [B_1,\ldots,B_d] \in \mathbb{R}^{n \times dK}$ denote the global spline
design matrix and let $Y = [Y_1,\ldots,Y_d]$ denote the matrix of centered
responses. Before search, we compute
\begin{align*}
G = B^\top B,
\,
H = B^\top Y,
\,
c_j = Y_j^\top Y_j,
\,\,\,\,
j \in \{1,\ldots,d\}.
\end{align*}
For a candidate parent set $S$, let $I(S)$ denote the columns of $B$
corresponding to the spline blocks of variables in $S$. The normal equations
for Equation~\eqref{eq:penalized-spline-fit} become
\begin{equation}
\left[
G_{I(S),I(S)} + P_S
\right]
\widehat{\beta}_{j,S}
=
H_{I(S),j}.
\label{eq:cached-normal-equations}
\end{equation}
Thus, each queried local regression reduces to extracting the cached blocks
$G_{I(S),I(S)}$ and $H_{I(S),j}$ and solving the corresponding penalized
linear system of dimension $|S|K$. A direct fit would instead have to form
$B_S^\top B_S$ and $B_S^\top Y_j$ from the observations for every query, with
sample-dependent costs proportional to $n(|S|K)^2$ and $n|S|K$, respectively.
For fixed bases and penalties, the cached formulation is equivalent to
the direct fit in Equation~\eqref{eq:penalized-spline-fit}, up to numerical
precision, but removes this repeated pass over the data.

% Thus, each local regression is evaluated by extracting cached blocks and
% solving a small penalized linear system. For fixed spline bases and penalties,
% this returns the same coefficient estimate as direct evaluation of
% Equation~\eqref{eq:penalized-spline-fit}, up to numerical precision, while
% avoiding repeated spline construction, Gram-matrix formation, and
% feature--response cross-products.

\paragraph{Residual likelihoods.}
After fitting the spline, the remaining step is to evaluate the residual
term in Equation~\eqref{eq:local-score}. We use two variants. The Gaussian
likelihood is computationally cheapest because it can be evaluated entirely
from cached quadratic quantities. The histogram likelihood is more flexible
because it estimates a univariate residual density and can better accommodate
non-Gaussian noise. Both variants use the same additive spline mean model and
the same BIC penalty; they differ only in the residual likelihood.

\noindent \emph{Gaussian likelihood.}
For the Gaussian variant, the cache suffices to compute the residual sum of
squares:
\begin{equation}
\operatorname{RSS}_{j,S}
=
c_j
-
2\widehat{\beta}_{j,S}^{\top}H_{I(S),j}
+
\widehat{\beta}_{j,S}^{\top}
G_{I(S),I(S)}
\widehat{\beta}_{j,S}.
\label{eq:cached-rss}
\end{equation}
This is exactly equal to
$\left\|Y_j - B_S\widehat{\beta}_{j,S}\right\|_2^2$, but does not require
materializing the residual vector. With the plug-in variance estimate
$
\widehat{\sigma}_{j,S}^2
=
\frac{\operatorname{RSS}_{j,S}}{n}
\label{eq:cached-gaussian-variance}
$%\end{equation}
, we obtain 
\begin{equation}
\mathcal{L}^{\mathrm{Gauss}}_j(S)
=
\frac{n}{2}
\left[
\log(2\pi)
+
1
+
\log\left(
\frac{\operatorname{RSS}_{j,S}}{n}
\right)
\right]
\label{eq:cached-gaussian-likelihood}
\end{equation}
for the Gaussian negative log-likelihood.
Thus, after preprocessing, Gaussian local-score evaluation depends on the
parent-set size but not on $n$.

\noindent \emph{Histogram likelihood.}
To allow more flexible residual distributions, we also use a histogram-based
negative log-likelihood. The histogram variant uses the same cached spline mean fit, but evaluates a
smoothed leave-one-out histogram density on the materialized residuals. This
gives a more flexible residual model at the cost of a sample-dependent
residual-scoring step. We give the exact binning, smoothing, and scoring rule
in Appx.~\ref{apx:implementation_details}.
% \begin{equation}
% \mathcal{L}^{\mathrm{Hist}}_j(S)
% =
% \operatorname{HistogramNLL}(\widehat{N}_{j,S}).
% \label{eq:histogram-likelihood}
% \end{equation}
This variant uses the same cached spline mean fit from
Equation~\eqref{eq:cached-normal-equations}, but must materialize
$\widehat{N}_{j,S}$ to estimate and evaluate the empirical residual density.
% It therefore retains a sample-size-dependent residual-scoring step, while still
% eliminating repeated mean-model fitting.

Algorithm~\ref{alg:spade-score} in Appx.~\ref{apx:cache} summarizes the procedure, and we provide implementation details in Appx.~\ref{apx:implementation_details}.
The preprocessing stage is executed once per dataset to enable fast queries during the combinatorial search.

\subsection{Search}
\label{subsec:search-topic}

The compiled score can be paired with any combinatorial search procedure that
queries decomposable local scores. We adopt the topological-order search
procedure introduced by \TOPIC, the strongest combinatorial baseline in
Section~\ref{sec:theory}. At each step, the procedure selects the next node in
the order, adds score-improving outgoing edges, and prunes redundant incoming
edges. It therefore repeatedly evaluates small candidate parent sets, for which
our cached spline score is particularly effective.

The published \TOPIC instantiation evaluated in
Section~\ref{sec:theory} fits a separate cubic-spline model for each queried
parent set, selecting its smoothing parameter during fitting. It also uses a
two-part MDL score with a constant coding cost per parameter and edge. We
instead use a shared fixed spline basis and smoothing penalty together with a
BIC-type criterion. The shared representation enables the efficient score evaluation through sufficient-statistic
compilation. The BIC penalty,
$k_S\log n$, provides increasingly strong complexity control as the sample
size grows, which is particularly relevant to the large-sample regimes.

Pairing the \TOPIC search procedure with our compiled spline score yields \ourmethod\footnote{\textbf{S}pline-\textbf{A}ccelerated
\textbf{D}AG \textbf{E}stimation}.
We refer to the Gaussian and histogram-based variants as
\ourmethodGauss and \ourmethodHist, respectively.
In Section~\ref{sec:exps}, we evaluate both variants extensively on a range of synthetic and 
real-world causal discovery benchmarks.

% \subsection{Search}
% \label{subsec:search-topic}

% Given an efficient decomposable score, we need a combinatorial search framework
% that can exploit repeated local-score evaluations. Natural candidates include
% the previously evaluated \GLOBE, \CAM, and \TOPIC. In the previous experiment,
% \TOPIC achieved the best performance in terms of both quality and runtime. We
% therefore use \TOPIC as the search backbone for our method. This choice is also
% computationally natural because \TOPIC is based on repeated sparse local-score
% evaluations.

% At a high level, \TOPIC interleaves ordering, edge addition, and pruning. At
% each step it selects the next node in the topological order, adds supported
% outgoing edges, and removes unsupported incoming edges. This differs from search
% strategies that separate graph construction from subsequent refinement. For
% example, \CAM first estimates a topological order and then prunes the dense DAG
% induced by that order using a separate variable-selection step. \GLOBE follows a
% different forward--backward greedy DAG search, adding candidate edges in a
% forward phase and deleting superfluous edges in a backward phase. Thus, while
% these methods differ in detail, they both separate constructive and corrective
% steps more explicitly than \TOPIC. In the \TOPIC-search, sparsity is instead maintained throughout
% the search rather than being imposed only after an initial constructive phase.

\begin{figure*}[t]
    \centering
\includegraphics[width=0.82\linewidth]{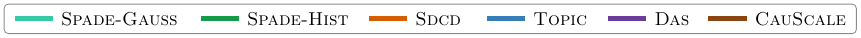}\par

    \begin{subfigure}{0.5\linewidth}
        \centering
        \includegraphics[width=0.5\linewidth]{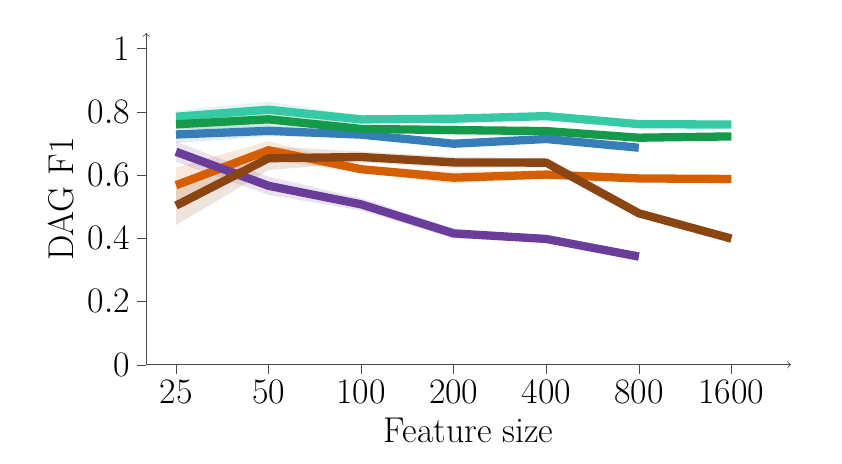}%
        \includegraphics[width=0.5\linewidth]{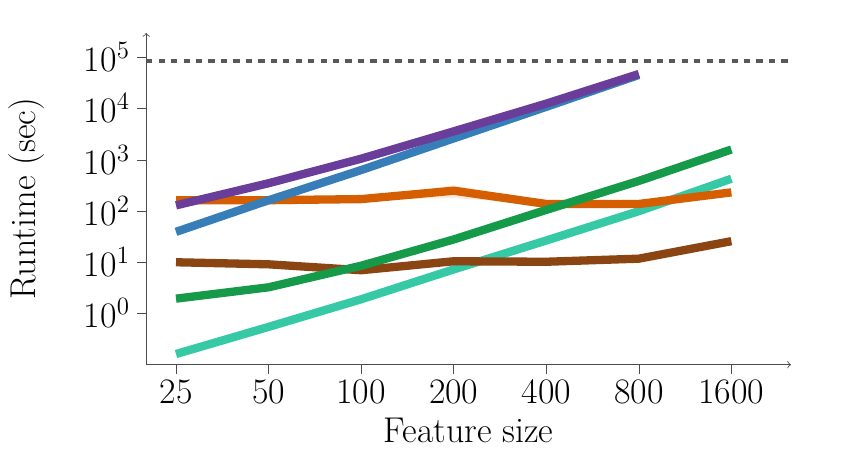}
        \caption{Feature complexity}
    \end{subfigure}%
    \begin{subfigure}{0.5\linewidth}
        \centering
        \includegraphics[width=0.5\linewidth]{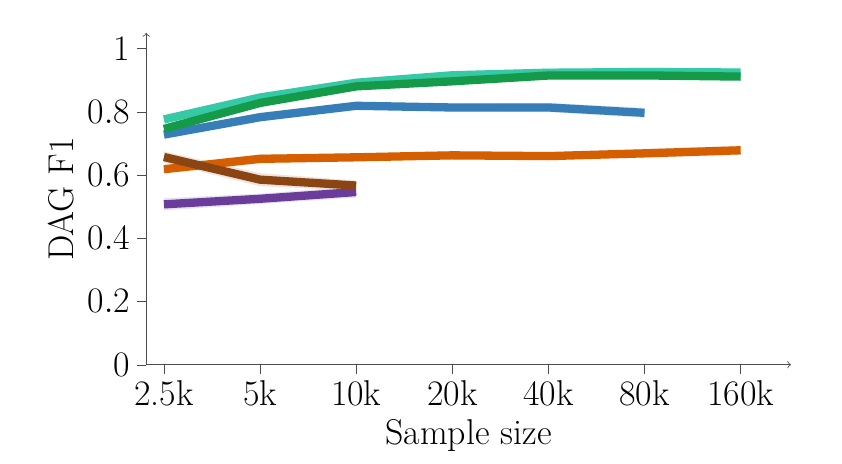}%
        \includegraphics[width=0.5\linewidth]{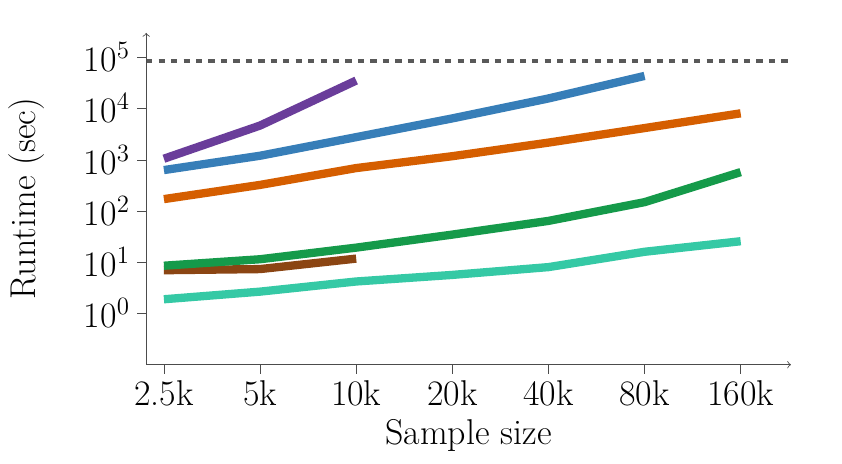}
        \caption{Sample complexity}
    \end{subfigure}
\caption{Scalability in features and samples on \emph{clean} data. Runtime and F1 score of the obtained DAG are shown in separate panels with matched dimensions. Results are averaged across 5 seeds, shaded regions correspond to standard errors.}
\label{fig:clean_scalability_spade}
\end{figure*}

\subsection{Complexity}
\label{subsec:complexity}
We analyze the cost of score evaluation and assume a fixed number $K$ of spline
basis functions per variable. The global spline design has $dK$ columns, so
constructing the cache costs $O(n(dK)^2)$ time and $O((dK)^2)$ storage, which
we write as $O(nd^2)$ time and $O(d^2)$ storage for fixed $K$.
\ourmethodHist additionally retains the spline design matrix in order to
construct residuals, requiring $O(ndK)=O(nd)$ storage.

Let $p = |S|$ be the size of a candidate parent set. Fixed-basis spline
evaluations from scratch require $O(np^2 + p^3)$ time: $O(np^2)$ to form local Gram and
feature--response products and $O(p^3)$ to solve the penalized system. 
\ourmethodGauss instead evaluates a local score in $O(p^2+p^3)=O(p^3)$ time by
using only extracted cache blocks and quadratic forms. \ourmethodHist uses the
same cached mean fit but additionally constructs and scores the residual vector,
adding $O(np+p^3)$ time per query. Thus, under bounded in-degree,
\ourmethodGauss requires constant time and \ourmethodHist requires $O(n)$ time
per local-score query.

Since \TOPIC performs $O(d^3)$ local-score queries under bounded degree \citep{xu:2025:topic}, the
total score-evaluation cost of \ourmethodGauss is
$O(nd^2+d^3)$, compared with $O(nd^3)$ for repeated direct spline evaluation.
For \ourmethodHist, the total asymptotic cost remains $O(nd^3)$ because each
query still scores residuals over the sample.
However, caching removes the repeated mean-fit computations, enabling substantial speed-ups
for both variants.

%% file: experiments.tex
\section{Experiments}
\label{sec:exps}

We evaluate \ourmethod along three complementary axes. First,
we evaluate \ourmethod on the scalability benchmark
from Sec.~\ref{sec:theory}. Second, at moderate
sample and feature sizes, we compare structural accuracy among a richer
set of competitors across a broad synthetic grid. Finally, we test whether the
observed gains transfer to real-world benchmark data. Additional experiments,
including no-cache ablations of \ourmethodGauss and \ourmethodHist, a scalability benchmark with heterogeneous mechanisms and noise distributions, and results
on the Sachs dataset~\citep{sachs:2005:causal}, are provided in
Appx.~\ref{apx:experiments}.
%In this section, we extensively evaluate \ourmethod on a large variety of synthetic datasets,  and a real-world benchmark. First, we evaluate \ourmethod in the clean setting of Sec.~\ref{sec:theory} to verify that \ourmethod improves the scalability--accuracy frontier as intended. Second, we repeat the experiment in a more realistic setting with varying mechanism classes and noise distributions to test whether the performance survives in more heterogenous data. Third, we do an extensive qualitative assessment of \ourmethod by comparing to a large set of competitors on 96 data generating settings and 480 corresponding synthetic datasets. Finally, we evaluate performance on the Causal Chamber benchmark \citep{gamella:2025:causalchamber}. In addition, in Appx.~\ref{apx:experiments}, we conduct an ablation study to isolate the cache-induced speed-up by running otherwise identical versions of \ourmethodHist and \ourmethodGauss that fit from scratch. We also evaluate performance on the small real-world Sachs dataset \citep{sachs:2005:causal}.

\subsection{Scalability}
\label{sec:scalability}
We evaluate scalability in the setting from Sec.~\ref{sec:theory}.
Fig.~\ref{fig:clean_scalability_spade} shows that both \ourmethod variants substantially advance the scalability--accuracy frontier. In the feature-scaling
experiment, they achieve higher F1 scores than \TOPIC while being orders of magnitud--
faster. The accuracy gain is not a consequence of caching, which only changes evaluation
cost; rather, it suggests that the BIC-penalized fixed-basis spline score gives
a more favorable parent-set criterion in this regime. 
At the largest feature size for which \TOPIC terminates, $d=800$,
\TOPIC requires 12.5 hours, whereas \ourmethodGauss finishes in 99.3 seconds ($456\times$ speed-up)
and \ourmethodHist in 386.9 seconds ($117\times$ speed-up). 
At $d=1600$, \TOPIC no longer completes within
24 hours, while \ourmethodGauss and \ourmethodHist terminate well within the limit. 
\SDCD and \CAUSCALE are faster at the largest feature sizes, but
achieve substantially lower F1 scores.

The advantage is even more pronounced in the sample-scaling experiment.
\ourmethodGauss and \ourmethodHist continue to improve with additional samples,
whereas \TOPIC plateaus. At the largest sample size for which \TOPIC
terminates, \ourmethodGauss and \ourmethodHist achieve speed-ups of
$2708\times$ and $291\times$, respectively. At $n=160{\,}000$, both variants
still complete, while maintaining the strongest structural accuracy. %among all methods.

To verify that these gains persist under more heterogenous mechanisms and noise distributions we repeated the experiment with a more diverse data-generating process in Appx.~\ref{apx:experiments}. The results show that the runtime advantage persists, and that \ourmethodHist is more robust to non-Gaussian noise than \ourmethodGauss.

\textbf{Overall}, these results show that \ourmethod advances the
empirical scalability--accuracy frontier on synthetic data. It improves the structural accuracy of the most accurate baseline while being orders of magnitude faster, handles both very large feature and sample sizes, and achieves
substantially higher F1 scores than the most scalable baselines.% \SDCD and \CAUSCALE.

\subsection{Robustness Across Data-Generating Regimes}
%We next assess whether the qualitative performance of \ourmethod is robust across data-generating regimes.

%In addition to the methods evaluated in Sec.~\ref{sec:theory}, we include \SCORE~\citep{rolland:2022:score}, \NOGAM~\citep{montagna:2023:nogam}, \CAPS~\citep{xu:2024:ordering}, and kernel-based \PC and \FCI~\citep{spirtes:1991:pc,spirtes:2001:fci,zhang:2012:kernel}. Since \PC and \FCI return Markov equivalence classes, we evaluate only their directed edges and treat ambiguous edges as abstentions.

%Fig.~\ref{fig:quality} shows \ourmethod and the three strongest competitors by average F1. The two variants of \ourmethod are complementary: \ourmethodHist achieves the strongest overall performance and is most robust under non-Gaussian noise, while \ourmethodGauss performs best in the Gaussian-noise regime and remains competitive elsewhere. Across the plotted conditions, one of the two variants achieves the best F1 score on every axis. We therefore recommend \ourmethodHist as the default unless scalability requirements are extreme or the residual noise is known to be close to Gaussian.

Previously, we have focused on runtime and F1 score as a function of feature and sample size.
We now assess robustness across graph density, mechanism class, and noise family
by reporting conditional F1 scores that marginalize over the remaining factors;
full regime-wise results are in Appx.~\ref{apx:metrics}.
The 96 settings cover ER DAGs with expected degrees in $\{2,3,4\}$, additive and non-additive mechanisms, and Gaussian, unimodal symmetric, unimodal asymmetric, or multimodal additive noise; details are given in Appx.~\ref{apx:data}.

To isolate robustness across data-generating regimes, and to compare to a broader set of competitors, we consider moderate problem sizes with $n\in\{1\,000,2\,000\}$, and $d \in \{20,50\}$. 
We extend the competitor set with \SCORE~\citep{rolland:2022:score}, \NOGAM~\citep{montagna:2023:nogam},
\CAPS~\citep{xu:2024:ordering}, and kernel-based \PC and
\FCI~\citep{spirtes:1991:pc,spirtes:2001:fci,zhang:2012:kernel}. Since \PC and
\FCI return Markov equivalence classes, we evaluate only their directed edges
and treat ambiguous edges as abstentions.

\begin{figure}[t]
    \includegraphics[width=\columnwidth]{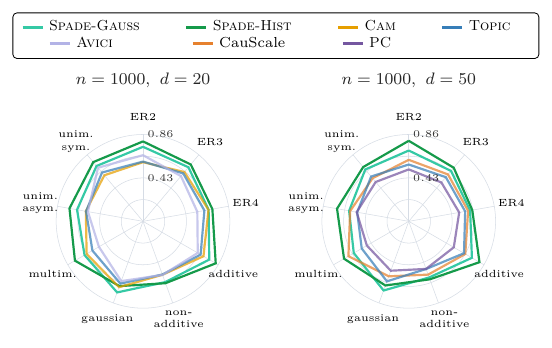}
\caption{Qualitative performance on synthetic datasets with \(n=1000\) in F1 score. Each spoke isolates the respective condition while marginalizing over remaining conditions.} %We present the two variants of \ourmethod, and the  3 strongest competitors.}
\label{fig:quality}
\end{figure}

We present the results for $n=1\,000$ in Fig.~\ref{fig:quality}, and defer the results for $n=2\,000$ to Appx.~\ref{apx:experiments}. We present \ourmethod and the three strongest competitors by
average F1 per problem size. Overall, \ourmethodHist achieves the
strongest performance and is most robust under non-Gaussian noise,
whereas \ourmethodGauss performs best in the Gaussian-noise regime and remains
competitive elsewhere. Across all marginalized settings, one of the two variants
obtained the best F1 score. 
We therefore recommend \ourmethodHist as the default unless scalability requirements are extreme or the residual noise is known to be close to Gaussian.

Finally, we aggregate F1 scores and runtimes across all synthetic settings to assess the moderate-size scalability--accuracy frontier. We include all methods except \PC and \FCI, which did not complete all $n=2000,d=50$ runs within 24 hours. In Fig.~\ref{fig:synthetic_pareto}, "Ours" refers to \ourmethodHist. Averaged across all settings, \ourmethodHist achieves the best F1 score, 0.748, compared to 0.616 for the strongest non-\ourmethod competitor, \TOPIC, while reducing mean runtime from 75.35 seconds to 1.24 seconds, a speed-up of roughly $61\times$. Even compared to the fastest external method, \CAUSCALE, which runs in 6.23 seconds on average but reaches only 0.542 F1, \ourmethodHist is both more accurate and about $5\times$ faster.

%In the next experiment, we test to what extent these improvements translate to real-world data.

\subsection{Real-World Data}
We finally evaluate whether the improvements on synthetic data transfer to real-world datasets. We focus on the \emph{Causal Chamber}
benchmark~\citep{gamella:2025:causalchamber}, a controlled physical benchmark
with known causal structure. 
As shown in Fig.~\ref{fig:cc_pareto}, \ourmethodHist achieves the strongest runtime--accuracy trade-off. It obtains an F1 score of 0.50 in 0.97 seconds on average. The closest competitor in F1 is \SDCD, with 0.49 F1, but it requires 520 seconds. The fastest competitor, \CAUSCALE, runs in 10 seconds but reaches only 0.27 F1.

We provide results on the Sachs protein-signaling dataset~\citep{sachs:2005:causal} in Appx.~\ref{apx:experiments}, where \ourmethod remains among the strongest and fastest methods.
Overall, the real-world results support the synthetic findings: \ourmethod is the only method in this comparison that combines competitive accuracy with consistently low runtime, advancing the empirical scalability--accuracy frontier beyond simulated benchmarks.
%     \centering

%     \begin{subfigure}{0.5\linewidth}
%         \centering
%         \includegraphics[width=\linewidth]{figs/tikz/causal_chamber_frontier.pdf}
%     \end{subfigure}%
%     \begin{subfigure}{0.5\linewidth}
%         \centering
%         \includegraphics[width=\linewidth]{figs/tikz/sachs_frontier.pdf}
%     \end{subfigure}
% \caption{Performance on the real-world datasets Causal Chamber (left) and Sachs (right). \ourmethod performs best both in terms of runtime and quality.}
% \end{figure}

%% file: conclusion.tex
\section{Conclusion}
\label{sec:conclusion}

We compared the scalability--accuracy trade-offs of four major families of nonlinear causal-discovery methods. Across the considered benchmarks, combinatorial methods provided the strongest signal, but were limited by repeated nonlinear local-score evaluations.

We introduced \ourmethod, a nonlinear additive score whose fixed-basis spline representation enables shared sample-dependent computations to be compiled and reused throughout sparse order search. This substantially advances the empirical scalability--accuracy frontier, retaining strong structural accuracy at problem sizes where existing nonlinear combinatorial methods become impractical. 

More broadly, our results highlight that scalability in nonlinear causal discovery should be evaluated along the full scalability--accuracy frontier and across all major algorithmic families. Such comparisons help localize where improvements occur and make the corresponding trade-offs explicit. We hope that our findings can guide future work both by providing a strong reference point and by revealing practical bottlenecks of existing approaches.

\subsection{Limitations \& Future Work}

Our comparison is empirical. The observed strengths and bottlenecks of the evaluated method families should therefore be interpreted as patterns in the studied benchmarks rather than universal characterizations of each approach. Reproducing these findings across further simulators and application domains is an important direction for future work.

We focus on causally sufficient, fully identifiable additive-noise models. Extending compiled nonlinear local scores to partially identifiable CPDAG settings, heteroscedastic mechanisms, latent confounding, and broader functional causal models remains open.

Finally, combinatorial search is not the only promising route toward scalable causal discovery. Improvements in score-derivative estimation, differentiable optimization, and amortized causal discovery could provide complementary solutions. %In particular, amortized methods offer attractive test-time efficiency, although robust extrapolation across graph sizes remains an important open challenge.

%% file: ai_use_statement.tex
\subsection*{AI use statement}

In this work, we used generative AI tools to assist with conceptual and
methodological discussion, technical formulation, and the interpretation and presentation of experimental
results. We additionally used generative AI tools to draft and edit text,
improve readability, and assist with coding. The authors reviewed
all AI-assisted material, checked technical statements against the underlying
derivations, implementation, and experimental results, and revised the final
manuscript accordingly. We take responsibility for the final content of this
work, including text, claims, code, or other artifacts produced with the aid of
generative AI.

%% file: checklist.tex
\section*{Checklist}

\begin{enumerate}

  \item For all models and algorithms presented, check if you include:
  \begin{enumerate}
    \item A clear description of the mathematical setting, assumptions, algorithm, and/or model. [Yes/No/Not Applicable]

    \textbf{Yes.} Sections~\ref{sec:prelim} and~\ref{sec:algo} describe the mathematical setting, assumptions, local score, compiled sufficient statistics, and search procedure.

    \item An analysis of the properties and complexity (time, space, sample size) of any algorithm. [Yes/No/Not Applicable]

    \textbf{Yes.} Section~\ref{subsec:complexity} analyzes the time and storage complexity of direct and compiled score evaluation and the resulting search procedure.

    \item (Optional) Anonymized source code, with specification of all dependencies, including external libraries. [Yes/No/Not Applicable]

    \textbf{Yes.} The supplementary material includes anonymized source code, a Dockerfile specifying the execution environment and dependencies, and commands for reproducing the experiments.
  \end{enumerate}

  \item For any theoretical claim, check if you include:
  \begin{enumerate}
    \item Statements of the full set of assumptions of all theoretical results. [Yes/No/Not Applicable]

    \textbf{Yes.} The assumptions used for the complexity analysis, including fixed spline-basis size and bounded parent-set size, are stated in Section~\ref{subsec:complexity}.

    \item Complete proofs of all theoretical results. [Yes/No/Not Applicable]

    \textbf{Not Applicable.} The paper does not present novel theorem or recovery-guarantee claims; its formal claims are direct computational-complexity derivations given in Section~\ref{subsec:complexity}.

    \item Clear explanations of any assumptions. [Yes/No/Not Applicable]

    \textbf{Yes.} The modeling assumptions are explained in Section~\ref{sec:prelim}, and the assumptions used in the complexity analysis are explained in Section~\ref{subsec:complexity}.
  \end{enumerate}

  \item For all figures and tables that present empirical results, check if you include:
  \begin{enumerate}
    \item The code, data, and instructions needed to reproduce the main experimental results (either in the supplemental material or as a URL). [Yes/No/Not Applicable]

    \textbf{Yes.} The supplementary reproduction package includes the experimental code, a containerized environment, exact paper presets, seeds, data-loading instructions, and commands for reproducing the main results.

    \item All the training details (e.g., data splits, hyperparameters, how they were chosen). [Yes/No/Not Applicable]

    \textbf{Yes.} Appendix~\ref{apx:data} gives the data-generation and sampling procedures, and Appendix~\ref{apx:implementation_details} reports the hyperparameters and how they were selected.

    \item A clear definition of the specific measure or statistics and error bars (e.g., with respect to the random seed after running experiments multiple times). [Yes/No/Not Applicable]

    \textbf{Yes.} Appendix~\ref{apx:metrics} defines the evaluation metrics and aggregation, while figure captions and the appendix report the number of seeds and identify shaded regions and tabulated uncertainties as standard errors.

    \item A description of the computing infrastructure used. (e.g., type of GPUs, internal cluster, or cloud provider). [Yes/No/Not Applicable]

    \textbf{Yes.} Appendix~\ref{apx:hardware} reports the compute-cluster setting, CPU model and allocation, and GPU model and memory.
  \end{enumerate}

  \item If you are using existing assets (e.g., code, data, models) or curating/releasing new assets, check if you include:
  \begin{enumerate}
    \item Citations of the creator If your work uses existing assets. [Yes/No/Not Applicable]

    \textbf{Yes.} The paper cites the creators of the existing datasets and baseline methods used in the evaluation.

    \item The license information of the assets, if applicable. [Yes/No/Not Applicable]

    \textbf{No.} The current manuscript does not list license information for the external datasets, code, and pretrained models used in the evaluation.

    \item New assets either in the supplemental material or as a URL, if applicable. [Yes/No/Not Applicable]

    \textbf{Yes.} The implementation introduced in this work is included in the supplementary reproduction package together with build and execution instructions.

    \item Information about consent from data providers/curators. [Yes/No/Not Applicable]

    \textbf{Not Applicable.} We use previously published benchmark datasets and do not collect data directly from participants or providers.

    \item Discussion of sensible content if applicable, e.g., personally identifiable information or offensive content. [Yes/No/Not Applicable]

    \textbf{Not Applicable.} The synthetic, physical-system, and protein-signaling benchmark data used here do not contain personally identifiable or offensive content.
  \end{enumerate}

  \item If you used crowdsourcing or conducted research with human subjects, check if you include:
  \begin{enumerate}
    \item The full text of instructions given to participants and screenshots. [Yes/No/Not Applicable]

    \textbf{Not Applicable.} This work did not use crowdsourcing or conduct research with human participants.

    \item Descriptions of potential participant risks, with links to Institutional Review Board (IRB) approvals if applicable. [Yes/No/Not Applicable]

    \textbf{Not Applicable.} This work did not use crowdsourcing or conduct research with human participants.

    \item The estimated hourly wage paid to participants and the total amount spent on participant compensation. [Yes/No/Not Applicable]

    \textbf{Not Applicable.} This work did not use crowdsourcing or conduct research with human participants.
  \end{enumerate}

\end{enumerate}

%% file: appendix.tex
\section{Appendix}
\label{sec:apx}

\subsection{Causal Discovery Methods}
\label{apx:methods}
This appendix describes the causal-discovery methods considered in the
empirical comparison.  The methods span combinatorial score optimization,
differentiable structure learning, score-matching and ordering-based discovery,
amortized causal discovery, and classical constraint-based references.

\paragraph{Combinatorial additive-score methods.}
\CAM estimates causal additive models by separating the problem into order
search and edge selection~\citep{buhlmann:2014:cam}.  It assumes an additive
structural equation model in which each variable is a sum of nonlinear effects
of its parents plus independent noise.  The method first searches for a
topological order using restricted likelihood scores for additive regressions
and then prunes the dense DAG induced by that order with sparse additive
regression.  This decoupling makes the search more manageable than enumerating
all DAGs, but each candidate still requires nonlinear regression fits.

\TOPIC is an information-theoretic causal-discovery method that also works in a
topological-order framework~\citep{xu:2025:topic}.  It scores candidate parent
sets with local code-length or likelihood criteria for continuous data and
interleaves ordering, edge insertion, and pruning rather than treating these as
fully separate stages.  The resulting search maintains a sparse graph throughout
the optimization and repeatedly compares nonlinear local mechanism scores,
which makes \TOPIC a strong but computationally demanding representative of
combinatorial nonlinear search.

\GLOBE searches directly over fully oriented causal networks using an
MDL-inspired score for functional causal models~\citep{mian:2021:globe}.  It
constructs a directed graph by greedily modifying parent sets and evaluating
whether the resulting local mechanisms yield a shorter description of the data.
Compared with order-only procedures, \GLOBE performs a more direct graph search,
but it relies on repeated local model evaluation and therefore faces the same
basic scalability pressure in high dimensions.

\paragraph{Differentiable structure learning.}
\NOTEARS replaces the discrete DAG constraint with a smooth acyclicity equality,
enabling gradient optimization over a continuous weighted adjacency matrix.
The nonlinear formulation parameterizes node mechanisms by neural networks and
uses sparsity of input derivatives to induce graph sparsity
~\citep{zheng:2020:learning}.  Thus, graph structure and nonlinear mechanisms
are learned jointly by continuous optimization, with acyclicity enforced through
the differentiable constraint.

\DAGMA is another continuous DAG-learning method, but replaces the matrix
exponential acyclicity function by a log-determinant characterization defined
over an M-matrix domain~\citep{bello:2022:dagma}.  The method optimizes a
sequence of unconstrained objectives along a barrier-like central path, with
sparsity regularization on the adjacency.  Its nonlinear variant combines this
acyclicity formulation with neural network mechanisms, yielding a continuous
graph--mechanism objective that is designed to have better numerical behavior
than earlier matrix-exponential relaxations.

\SDCD is designed to stabilize differentiable causal discovery at larger
dimensions~\citep{nazaret:2024:stable}.  It uses an acyclicity constraint with
quadratic computational structure and a training procedure tailored to sparse
graphs, avoiding the cubic matrix operations that dominate several earlier
continuous relaxations.  The method separates training into stages that first
obtain a stable sparse structural representation and then refine the graph and
mechanisms, making it a scalable representative of differentiable DAG learning.

\paragraph{Score-matching and ordering methods.}
\SCORE recovers a causal order in nonlinear additive Gaussian-noise models from
derivatives of the observational log density~\citep{rolland:2022:score}.  The
key signal is based on the score function
\(\nabla_x \log p(x)\) and its derivatives: leaf nodes can be identified from
properties of the Hessian of the log density, removed, and the procedure
iterated.  After an order has been estimated, sparse edge selection can be
performed within the DAGs compatible with that order.

\NOGAM extends the score-matching route beyond Gaussian noise by using
additive-model assumptions with arbitrary noise distributions~\citep{montagna:2023:nogam}.
It retains the idea of estimating density-score-derived quantities to infer an
order, but avoids relying on the Gaussian-noise form of the original \SCORE
criterion.  This makes \NOGAM a score-matching method for nonlinear additive
models with weaker restrictions on the exogenous noise distribution.

\DAS uses score matching to recover the whole graph in nonlinear additive
Gaussian-noise models while reducing the cost of the pruning stage
substantially~\citep{montagna:2023:das}.  Instead of first estimating a full
order and then applying an expensive dense pruning pass, it exploits
second-derivative information of the log density to remove many candidate
edges more directly.  The method therefore targets the same score-derived
identifiability signal as \SCORE, but couples order estimation with a more
efficient graph-recovery procedure.

\DIFFAN is an order-estimation method that replaces kernel or Stein estimates
of the required Hessian quantities by a diffusion-model estimator
~\citep{sanchez:2023:diffan}.  It learns a neural score model and uses the
estimated Hessian information to identify leaves recursively, yielding a
topological order for an additive-noise model.  The central idea is to replace
explicit high-dimensional density-derivative estimation by a learned score
model whose derivatives can be used for causal ordering.

\SCINO also targets ordering-based causal discovery from score-derived Hessian
information, but uses a score-informed neural operator to approximate the
Hessian diagonal more stably than directly differentiating a score network
~\citep{kang:2026:scino}.  It represents score-related functions in a smooth
operator-learning framework, aiming to preserve the ordering information needed
for additive-noise causal discovery while improving numerical stability over
direct Hessian estimation.

\CAPS is an ordering-based method for data containing both linear and nonlinear
causal relations~\citep{xu:2024:ordering}.  It estimates an order from
score-derived Hessian information and then uses parent scores to guide
post-processing: weak candidate parents are pruned before additional high-score
edges are considered subject to acyclicity.  The method is designed for mixed
linear and nonlinear mechanisms, where a single purely linear or purely
nonlinear criterion may be poorly matched to the data-generating process.

\paragraph{Amortized causal discovery.}
\AVICI performs amortized variational inference for causal structure learning
~\citep{lorch:2022:avici}.  A neural inference model is pretrained on simulated
SCMs and then maps a new dataset directly to edge probabilities, so test-time
inference is a forward pass rather than dataset-specific graph search.  Its
architecture is designed to respect permutation symmetries of the causal
structure learning problem, while its statistical behavior is governed by the
synthetic simulator distribution used during pretraining.

\CAUSCALE is a neural causal-discovery foundation model designed for large
graphs~\citep{peng:2026:causcale}.  It improves the memory and runtime profile
of transformer-style amortized discovery by compressing data embeddings and
using tied attention weights, while combining a data stream with a graph-prior
stream.  As with other amortized methods, the computational cost of learning is
shifted to pretraining, after which structure prediction for a new dataset is
performed by neural inference.

\paragraph{Constraint-based references.}
\PC and \FCI are included as classical constraint-based references in the
qualitative synthetic comparison~\citep{spirtes:1991:pc,spirtes:2001:fci}.
\PC assumes causal sufficiency and recovers a Markov equivalence class by
iteratively testing conditional independences, whereas \FCI allows latent
confounding and selection effects and returns a partial ancestral graph.
Kernel conditional independence tests extend these procedures to nonlinear
relationships~\citep{zhang:2012:kernel}.  Unlike the additive-noise methods
above, constraint-based methods identify an equivalence class under Markov and
faithfulness assumptions rather than a single fully oriented DAG in general.

\subsection{Data}
\label{apx:data}
\paragraph{Synthetic data}
All synthetic datasets are generated from causally sufficient structural causal
models with independent additive noise.  Unless stated otherwise, we first draw
a random topological order \(\pi\) on \(d\) variables and then sample an
Erd\H{o}s--R'enyi DAG by adding each order-compatible edge independently with
probability
\[
    p = \min\{ \bar{k}/(d-1), 1\},
\]
where \(\bar{k}\) is the target expected degree.  If the resulting graph is
empty, we add the edge from the first to the last variable in the sampled order.
Each present edge receives a signed weight whose magnitude is sampled uniformly
from \([0.5,2]\).  Variables are generated in topological order.  Root variables
are exogenous noise variables.  For a non-root node \(j\), with parent values
\(X_{\Pa(j)}\), we sample
\[
    X_j = f_j(X_{\Pa(j)}) + N_j .
\]
The noise variables \(N_j\) are independent across nodes and samples.  After
generation, every observed variable is standardized to zero empirical mean and
unit empirical variance.

The generic simulator supports four mechanism classes.  The linear class uses
the sampled edge weights directly.  The additive class generates a sum of
univariate nonlinear parent effects.  For each non-root node, the nonlinear
effect family is sampled from basis functions, polynomials, random Fourier
features approximating a Gaussian process, and one-hidden-layer neural networks.
Basis effects use random transformations from
\(\{\sin,\cos,\tanh,\mathrm{sigmoid},x^3,\mathrm{sign}(x)\sqrt{|x|},
\exp(-|x|)\}\) with random scale and shift; polynomial effects use clipped
powers of degree \(2\) to \(4\); Gaussian-process effects use \(16\) to \(48\)
random Fourier features with lengthscale in \([0.4,2.5]\); neural mechanisms
use hidden width \(4\) to \(12\) and an activation sampled from
\(\{\tanh,\mathrm{ReLU},\mathrm{sigmoid},\sin\}\).  The nonadditive class adds
parent-parent interaction terms to the additive effects; for Gaussian-process
and neural-network generators it also includes a multivariate parent effect.
The never-additive class removes the additive component whenever a node has at
least two parents, leaving only interaction effects; nodes with a single parent
fall back to a univariate nonlinear effect.  In heterogeneous settings, the
mechanism class is sampled independently for each non-root node from the
available classes.

Noise distributions are sampled from four families.  The Gaussian family is
standard normal.  The unimodal symmetric family contains Laplace noise scaled
to unit variance, Student-\(t\) noise with three degrees of freedom scaled by
\(\sqrt{3}\), and uniform noise on \([-\sqrt{3},\sqrt{3}]\).  The unimodal
asymmetric family contains centered exponential noise, standardized lognormal
noise, standardized Gamma noise, and a standardized skew-normal-like mixture
formed by adding an absolute Gaussian component.  The multimodal family
contains two- and three-component Gaussian mixtures, three-component Laplace
and Student mixtures, and a four-component mixed Gaussian/Laplace/Student/
exponential mixture; all multimodal noises are standardized.  In homoscedastic
settings, each non-root node receives a constant noise scale sampled uniformly
from \([0.4,1]\), except in the controlled scalability experiments described
below.

\emph{Controlled scalability.}
The controlled scalability experiments use a deliberately clean additive-noise
setting.  The graph is sampled with exactly
\(\operatorname{round}(\bar{k}d/2)\) order-compatible edges rather than only in
expectation, with \(\bar{k}=3\).  All non-root structural mechanisms are
Gaussian-process functions represented by random Fourier features with
lengthscale in \([0.4,2.5]\), and all noise variables are Gaussian.  Non-root
noise scales are fixed to \(1\), so the only scale variation is introduced by the
final column standardization.  The feature-scaling panel fixes
\(n=2{,}500\) and varies
\[
    d \in \{25,50,100,200,400,800,1\,600\},
\]
whereas the sample-scaling panel fixes \(d=100\) and varies
\[
    n \in \{2{\,}500,5{\,}000,10{\,}000,20{\,}000,40{\,}000,80{\,}000,160{\,}000\}.
\]
Each setting is averaged over five independently generated datasets.  The
runtime-ablation experiment uses the same controlled data-generating process and
the same feature- and sample-scaling grids.

\emph{Heterogeneous scalability.}
The heterogeneous scalability experiment keeps the same graph-density and
problem-size grids as the controlled scalability experiment, but replaces the
clean GP/Gaussian mechanisms by node-wise heterogeneous mechanisms and noises.
The graph is an order-based Erd\H{o}s--R'enyi DAG with expected degree
\(\bar{k}=3\).  For each non-root node, the mechanism class is sampled at
random, and nonlinear mechanisms further sample their concrete generator from
the basis, polynomial, Gaussian-process, and neural-network families described
above.  The noise distribution is also sampled node-wise by first choosing one
of the Gaussian, unimodal symmetric, unimodal asymmetric, or multimodal noise
families and then sampling a concrete distribution from that family.  The data
are homoscedastic, with node-wise constant noise scales sampled from
\([0.4,1]\), and are standardized column-wise after generation.  We use five
independent seeds per setting.

\emph{Qualitative synthetic comparison.}
The qualitative comparison uses the same generic SCM simulator, but varies the
data-generating factors in a full moderate-size grid.  We use
\[
    (n,d) \in \{(1000,20),(2000,20),(1000,50),(2000,50)\},
    \qquad
    \bar{k}\in\{2,3,4\},
\]
with either additive or never-additive mechanisms.  The noise family is one of
Gaussian, unimodal symmetric, unimodal asymmetric, and multimodal noise.  This
yields \(4 \times 3 \times 2 \times 4 = 96\) data-generating settings.  For
each setting we generate five independent datasets, giving \(480\) synthetic
datasets in total.  The noise family is resolved independently for each node to
one of the concrete distributions in that family, and all datasets are
homoscedastic and column-standardized.

\emph{Differentiable-mechanism diagnostic.}
The diagnostic experiment in Appx.~\ref{apx:experiments} uses a separate
non-additive neural SCM designed to be favorable to neural mechanism learners.
For each \(d\in\{10,25,50,100\}\) and seed \(s\in\{0,\ldots,4\}\), we sample
\(n=1000\) observations from an order-based Erd\H{o}s--R'enyi DAG with expected
degree \(4\), forcing one edge if the sampled graph is empty.  Root variables
are independent standard Gaussian noises.  For each non-root variable, the main
effect is a random one-hidden-layer neural network of its parents with hidden
width sampled uniformly from \(\{6,\ldots,16\}\), Gaussian weights scaled by the
input width, Gaussian biases, and activation sampled from
\(\{\tanh,\mathrm{ReLU},\mathrm{sigmoid},\sin\}\).  If a node has at least two
parents, every parent pair contributes an additional random neural interaction,
multiplied by a coefficient sampled uniformly from \([0.5,1.5]\).  The
deterministic effect is standardized within the generated sample and then
combined with Gaussian noise scaled by a coefficient sampled uniformly from
\([0.4,1]\).  The final observed variables are standardized column-wise.

\paragraph{Real-world data}
The real-world experiments use two observational tabular benchmarks with
reference graphs.  For Causal Chamber~\citep{gamella:2025:causalchamber}, we
use the light-tunnel dataset \texttt{lt\_interventions\_standard\_v1} and its
\texttt{uniform\_reference} experiment, which provides non-image observations
from the standard light-tunnel configuration together with a chamber-level
ground-truth graph.  We keep the numeric measurement variables, remove rows with
missing or non-finite values, discard constant columns, and standardize each
variable.  The reference graph is restricted to the retained variables.  The
reported Causal Chamber setting contains \(10{,}000\) observations of \(23\)
variables and \(39\) reference edges; the table reports five bootstrap
resamples of this observational table.

For Sachs~\citep{sachs:2005:causal}, we use the single-condition observational
Benchpress \citep{rios:2025:benchpress} version of the protein-signaling data.   The reference graph is the standard 11-node consensus DAG over
the measured signaling proteins.  After removing one incomplete row, the reported
Sachs setting contains five bootstrap samples of \(853\) observations, \(11\)
variables, and \(17\) reference edges.

\subsection{Hardware}
\label{apx:hardware}
Experiments were run on a compute cluster. GPU-based methods were allocated one
NVIDIA A100 40\,GB GPU and four CPU threads per run. CPU-based methods were
run on nodes equipped with AMD EPYC 7773X processors; each CPU run was allocated
four CPU threads and no GPU. Reported runtimes are wall-clock runtimes of the
benchmark invocation.

\subsection{Experiments}
\label{apx:experiments}
In this section we present additional experimental details for the experiments in the main paper and provide additional experiments.

\begin{figure*}[t]
    \centering
\includegraphics[width=0.82\linewidth]{figs/tikz/spade_scalability_methods_legend.pdf}\par

    \begin{subfigure}{0.5\linewidth}
        \centering
        \includegraphics[width=0.5\linewidth]{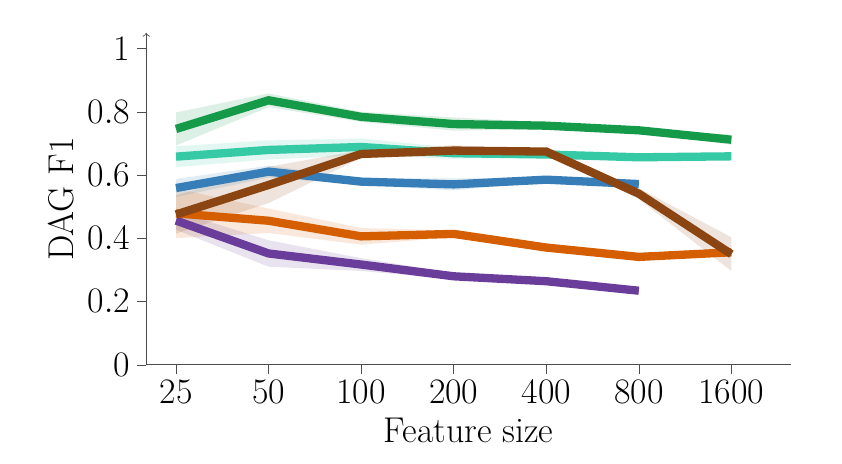}%
        \includegraphics[width=0.5\linewidth]{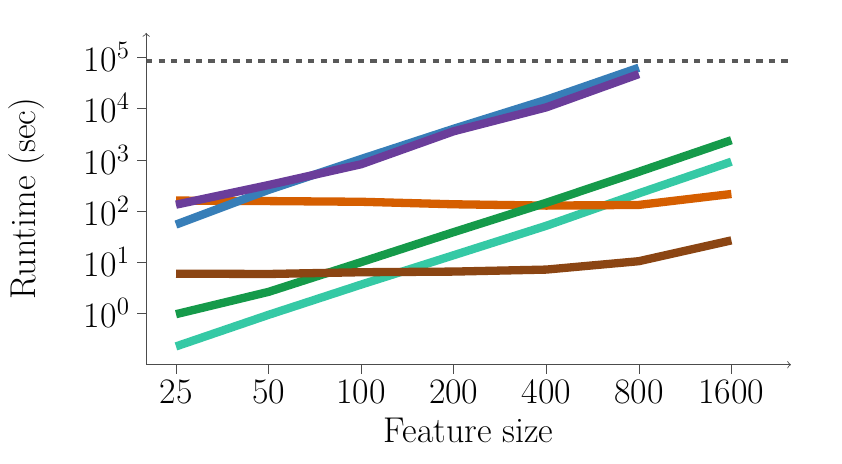}
        \caption{Feature complexity.}
    \end{subfigure}%
    \begin{subfigure}{0.5\linewidth}
        \centering
        \includegraphics[width=0.5\linewidth]{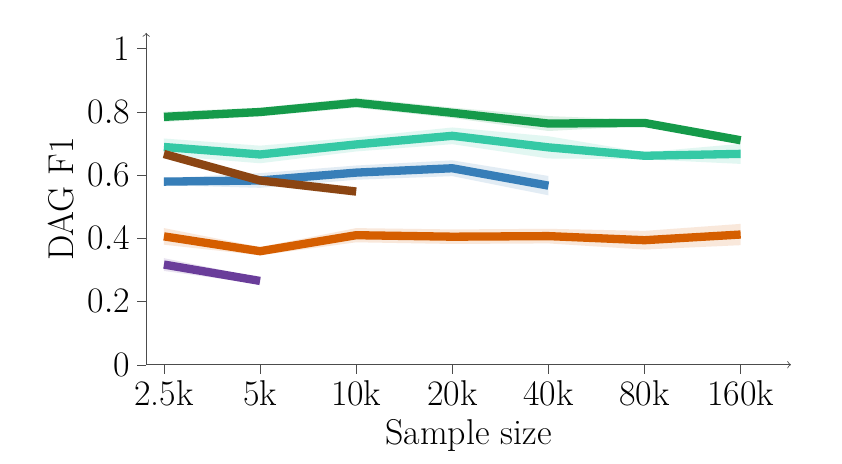}%
        \includegraphics[width=0.5\linewidth]{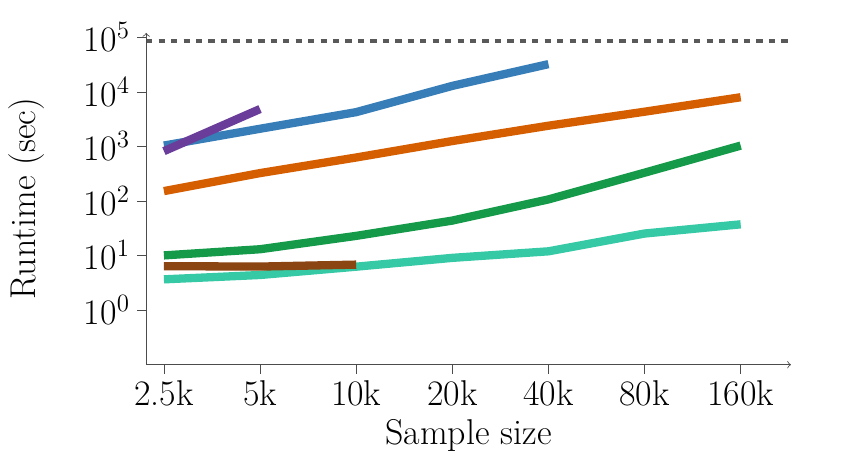}

        \caption{Sample complexity.}
    \end{subfigure}

\caption{Scalability in features and samples on \emph{heterogeneous} data. Runtime and F1 score of the obtained DAG are shown in separate panels with matched dimensions. Results are averaged across 5 seeds, shaded regions correspond to standard errors.}
\label{fig:scalability_hetero}
\end{figure*}

\paragraph{Heterogeneous benchmark.}
To verify that the scalability--accuracy gains observed in Sec.~\ref{sec:scalability} persist under more heterogeneous mechanisms and noise distributions, we run a variant of the controlled scalability experiment with node-wise heterogeneous mechanisms and noises, we repeat the experiments from Sec.~\ref{sec:scalability} with the same feature- and sample-scaling grids, but with node-wise heterogeneous mechanisms and noise distributions.
The diverse simulator samples node-wise mechanism classes and
noise families, including linear, additive, non-additive, and interaction-based
nonlinear mechanisms, as well as Gaussian, unimodal, asymmetric, and multimodal
additive noise. We describe the data generation process in detail in Appx.~\ref{apx:data}.

Fig.~\ref{fig:scalability_hetero} shows that the gains persist in the diverse setting.
In the feature-scaling experiment, both variants again
outperform \TOPIC in F1 score. \ourmethodHist achieves the strongest accuracy,
consistent with its more flexible residual likelihood. At $d=1600$, \TOPIC does
not complete within 24 hours, while \ourmethodGauss and \ourmethodHist finish
in 918 and 2416 seconds. As before, \SDCD and \CAUSCALE remain faster at the largest
feature sizes, but are substantially less accurate.
The sample-scaling experiment further highlights the robustness of the
histogram score. \ourmethodHist reaches an F1 score of 0.829, compared to
0.724 for \ourmethodGauss and 0.622 for \TOPIC. 
In terms of runtime, \ourmethodGauss is even faster than \SDCD and \CAUSCALE, which runs out of memory after $n=10{\,}000$ samples. 
\paragraph{Differentiable Structure Learning Experiment}

To investigate the lower structural quality of the differentiable methods compared to \TOPIC observed in Sec.~\ref{sec:theory}, we test whether the learned mechanisms are capable of guiding the search towards the true DAG. While these mechanisms are typically more expressive than the additive spline models used by combinatorial methods, the joint optimization of a DAG parametrization and nonlinear mechanisms can get stuck in local optima: the mechanisms may adapt to the currently preferred candidate DAG and fail to assign better scores to better structures.

To test this hypothesis, we consider \NOTEARS, \DAGMA and \SDCD. We use a data-generating setup that is favorable to neural mechanism learners. For each dimension \(d\in\{10,25,50,100\}\) and seed \(s\in\{0,\ldots,4\}\), we sample \(n=1000\) observations from a random nonlinear additive-noise SCM. The graph is an Erdos--Renyi DAG generated by first sampling a random topological order and then adding each forward edge independently with probability \(4/(d-1)\), corresponding to expected degree \(4\); if this produces an empty graph, one forward edge is added. Root variables are independent standard Gaussian noise variables. For non-root variables, the structural function is a random one-hidden-layer neural network of its parents, with hidden width sampled uniformly from \(\{6,\ldots,16\}\), Gaussian weights scaled by the input width, Gaussian biases, and an activation sampled from \(\{\tanh,\mathrm{ReLU},\mathrm{sigmoid},\sin\}\). To make the mechanisms non-additive, whenever a node has at least two parents we additionally add random neural effects of every parent pair, each multiplied by a coefficient sampled uniformly from \([0.5,1.5]\). The deterministic effect is standardized within the generated sample and then combined with Gaussian noise scaled by a coefficient sampled uniformly from \([0.4,1.0]\). Finally, all variables are standardized column-wise before fitting the methods.

During the training of each method, we freeze the learned mechanisms at checkpoints and test how well they rank the true DAG compared to \(1000\) dense DAGs induced by independent random orders. Each candidate DAG is scored by applying its mask to the frozen learned mechanism and evaluating the resulting reconstruction or likelihood score on the validation split. We compare these rankings to an additive regression spline score, which is misspecified for the non-additive MLP mechanisms but refits a separate regression for each candidate DAG.

\begin{figure*}
    \includegraphics[width=\linewidth]{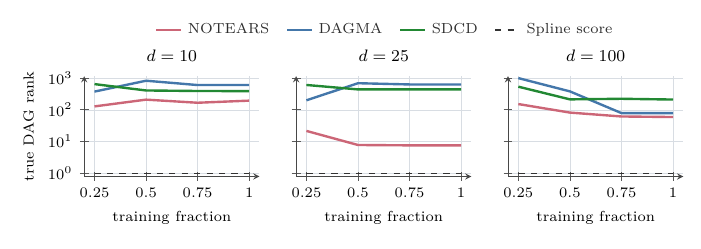}
    \caption{The mechanisms learned by differentiable structure learning methods do not rank the true DAG the highest reliably.}
    \label{fig:diff_fail}
\end{figure*}

We illustrate the results in Fig.~\ref{fig:diff_fail}. No differentiable method reliably ranks the true DAG highest among only \(1000\) random DAGs. No differentiable method achieves an ideal rank. In contrast, the spline score ranks the true DAG highest consistently despite the generating mechanisms being non-additive. This is possible because ranking the true DAG highest among only \(1000\) random-order DAGs is a weak requirement: most random orders are far from the true one. The fact that the learned mechanisms fail this task suggests that the benchmarked differentiable learners get stuck in poor local optima of the joint DAG--mechanism space. Bridging the gap to the structural quality of combinatorial methods therefore likely requires optimization strategies that produce more reliable structural ranking signals and escape these local optima.

\paragraph{Runtime Ablation}

\begin{figure*}[t]
    \centering
    \includegraphics[width=0.78\linewidth]{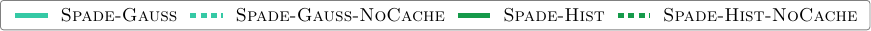}\par
    \begin{subfigure}{0.5\linewidth}
        \centering
        \includegraphics[width=\linewidth]{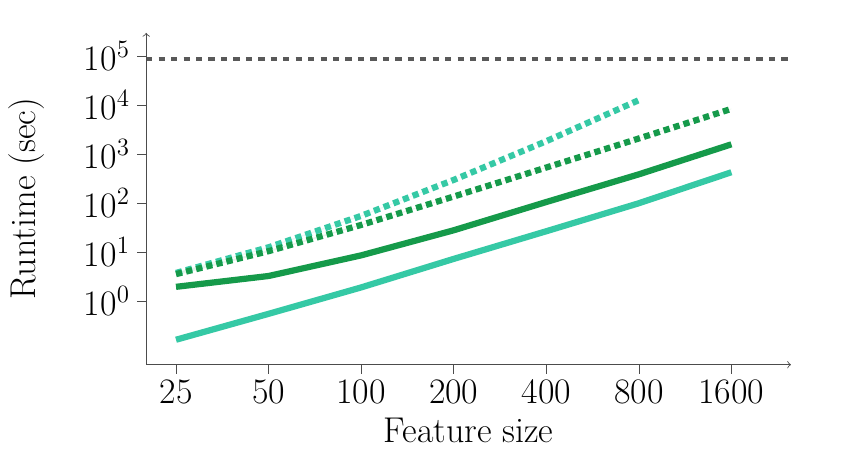}
        \caption{Feature complexity.}
    \end{subfigure}%
    \begin{subfigure}{0.5\linewidth}
        \centering
        \includegraphics[width=\linewidth]{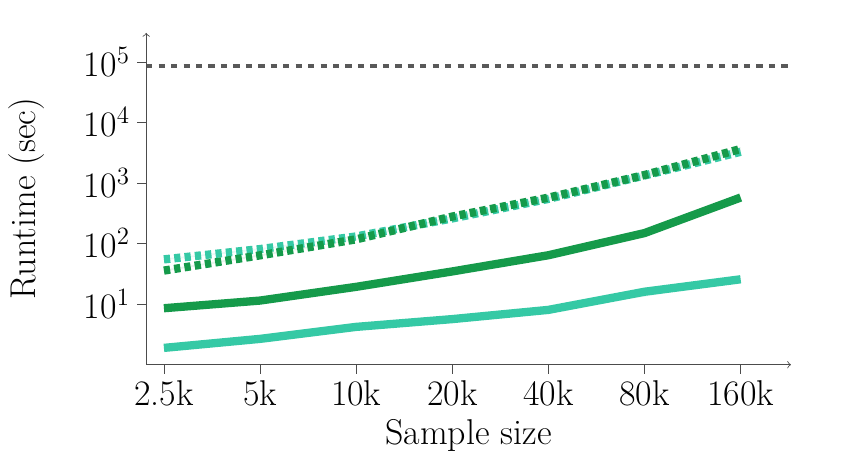}
        \caption{Sample complexity.}
    \end{subfigure}
    \caption{Runtime ablation of the cached sufficient-statistics implementation. Each cached variant is compared to an otherwise identical variant that refits the additive-spline score from scratch. Results are averaged across 5 seeds, and shaded regions show standard errors.}
    \label{fig:runtime-variants}
\end{figure*}

To isolate the gains of compiling the Gram matrices and feature--response
cross-products, we run \ourmethodGauss and \ourmethodHist without compilation on setting with Gaussian processes and Gaussian noise. We refer to these variants as  \ourmethodGauss-\textsc{NoCache} and \ourmethodHist-\textsc{NoCache}. We present the results in Fig.~\ref{fig:runtime-variants}. As expected we observe significant speed-up through caching, especially for \ourmethodGauss.

\label{app:runtime-variants}

\begin{figure}
    \centering
    \includegraphics[width=0.5\textwidth]{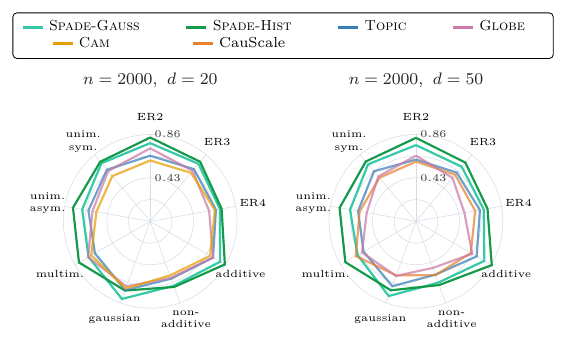}
\caption{Qualitative performance on synthetic datasets with \(n=2000\) in F1 score. Each spoke isolates the respective condition while marginalizing over remaining conditions.} %We present the two variants of \ourmethod, and the  3 strongest competitors.}
\label{fig:quality_2000}
\end{figure}

\paragraph{Qualitative results at $n=2000$}
We present the qualitative synthetic comparison at $n=2000$ in Fig.~\ref{fig:quality_2000}. While the set of the three strongest competitors differs, the patterns of the \ourmethod-variants are similar to the $n=1000$-setting, which is presented in the main paper.

\paragraph{Sachs}

We evaluate the two variants of \ourmethod and all competitors on the real-world Sachs dataset \citep{sachs:2005:causal}.
Because of the skewedness of the distributions, we applied a $\log$ and $\operatorname{arcsinh}$ transformations to the data. We report the results for the raw data and the two transformations in Tab.~\ref{tab:sachs_observational_transform_benchmark}. While the \ourmethod-variants do not achieve the best F1 score, they are among the 3 highest F1 scores in all 3 settings, and consistently achieve the lowest SHD, while also achieving the lowest runtimes.

\subsection{Cached Sufficient Statistics}

We present the algorithm that summarizes the algebraic compilation algorithm underlying \ourmethod in Algorithm~\ref{alg:spade-score}.

\label{apx:cache}
\begin{algorithm}[t]
\caption{Compiled local-score oracle used by \ourmethod}
\label{alg:spade-score}
\begin{algorithmic}[1]
\REQUIRE Dataset $X \in \mathbb{R}^{n \times d}$; residual likelihood $\ell$;
spline basis size $K$;
spline-penalty rule $S \mapsto P_S$
\ENSURE Gaussian cache $\mathcal{C}_{\mathrm{Gauss}} = (G,H,c)$ or histogram
cache $\mathcal{C}_{\mathrm{Hist}} = (B,G,H,c)$

\STATE Construct centered spline blocks $B_1,\ldots,B_d$ and centered
responses $Y_1,\ldots,Y_d$
\STATE $B \gets [B_1,\ldots,B_d]$
\STATE $G \gets B^\top B$, \quad $H \gets B^\top Y$
\STATE $c_j \gets Y_j^\top Y_j$ for all $j \in V$
\IF{$\ell = \textsc{Gaussian}$}
    \STATE Store $\mathcal{C}_{\mathrm{Gauss}} \gets (G,H,c)$
\ELSE
    \STATE Store $\mathcal{C}_{\mathrm{Hist}} \gets (B,G,H,c)$
\ENDIF

\STATE \textbf{LocalScore}$(j,S,\ell,\mathcal{C})$
\STATE $I \gets I(S)$
\STATE $A_S \gets G_{I,I} + P_S$, \quad $h_{j,S} \gets H_{I,j}$
\STATE Solve $A_S\widehat{\beta}_{j,S} = h_{j,S}$

\IF{$\ell = \textsc{Gaussian}$}
    \STATE $\operatorname{RSS}_{j,S} \gets
    c_j - 2\widehat{\beta}_{j,S}^{\top}h_{j,S}
    + \widehat{\beta}_{j,S}^{\top}G_{I,I}\widehat{\beta}_{j,S}$
    \STATE \RETURN $\mathcal{L}^{\mathrm{Gauss}}_j(S)
    + \operatorname{pen}_{\mathrm{BIC}}(j,S)$
\ELSE
    \STATE $\widehat{N}_{j,S} \gets
    Y_j - B_S\widehat{\beta}_{j,S}$
    \STATE \RETURN $\operatorname{HistogramNLL}(\widehat{N}_{j,S})
    + \operatorname{pen}_{\mathrm{BIC}}(j,S)$
\ENDIF
\end{algorithmic}
\end{algorithm}

\subsection{Additional Metrics}
\label{apx:metrics}
In this appendix, we present detailed results for the experiments in the main
paper, including additional graph-recovery and runtime metrics.  The controlled
tables report directed edge precision, recall, and F1 by comparing the ordered
edge sets of the learned graph and the ground-truth DAG.  Skeleton metrics use
the same precision, recall, and F1 definitions after replacing each directed
graph by its undirected skeleton.  Empty predicted graphs have precision \(1\)
only when the true graph is also empty, and recall \(1\) only when both graphs
are empty.  We report structural Hamming distance (SHD) with a reversed edge
counting as one operation: if \(M\) is the number of missing directed edges,
\(E\) the number of extra directed edges, and \(R\) the number of reversed true
edges, then \(\mathrm{SHD}=M+E-R\).  The normalized topological divergence is
the fraction of true directed edges whose endpoints appear in the wrong order in
the estimated topological order.  Runtime is measured in seconds for each
method run.  The table column \(n\) reports the number of successful
rows entering each mean, and failed, timed-out, or OOM runs are excluded from
the corresponding means.  Entries are reported as means with standard errors
over these successful rows.  The controlled
feature-scaling results are shown in
Tables~\ref{tab:controlled_scalability_feature_size_n_features_25}--\ref{tab:controlled_scalability_feature_size_n_features_1600},
and the controlled sample-scaling results are shown in
Tables~\ref{tab:controlled_scalability_sample_size_n_samples_2500}--\ref{tab:controlled_scalability_sample_size_n_samples_160000}.
The corresponding heterogeneous scalability results are reported in
Tables~\ref{tab:heterogeneous_scalability_feature_size} and
\ref{tab:heterogeneous_scalability_sample_size}.  
The Causal Chamber results are shown in
Table~\ref{tab:real_world_benchmark}.

\input{tables/generated/appendix_experiment_tables.tex}
\subsection{Implementation Details}
\label{apx:implementation_details}
We used the default or author-provided hyperparameters of the corresponding
implementations wherever possible.  This choice is deliberate: in causal
discovery, tuning method-specific hyperparameters without access to the
ground-truth graph is not straightforward, and using fixed defaults avoids
giving methods different amounts of oracle information.  We did not tune
hyperparameters against the benchmark ground truth; computational batch sizes
and subsampling parameters were adjusted only when needed to make the methods
fit GPU memory. 

For methods with stochastic optimization, we ran one predetermined seed per method–dataset pair. We did not use random restarts or select the best run across seeds. The seed policy and exact seeds are provided in the released code.

\paragraph{\ourmethod.}
To ensure a fair comparison, we fixed sensible defaults for all hyperparameters before evaluation, and did not perform any hyperparameter tuning.
Both variants of \ourmethod use the same additive univariate spline regression
for local parent-set scores.  The spline basis is the
\texttt{SplineTransformer} basis from scikit-learn: cubic B-splines
(\texttt{degree=3}), no bias column (\texttt{include\_bias=False}), and linear
extrapolation outside the fitted knot range.  The implementation uses an
internal \texttt{n\_splines=20} setting, which is converted to
\texttt{n\_knots=18} for scikit-learn; with cubic splines and
\texttt{include\_bias=False}, this gives \(19\) output basis columns per
univariate parent block.  The resulting basis columns are centered before
scoring.  The local regression uses ridge regularization \(10^{-6}\) and a
second-difference smoothing penalty with weight \(0.5\).
Scores are evaluated on the deviance scale with a BIC
penalty.  For a candidate parent set \(S\), \ourmethodGauss uses the local score
\[
  \mathcal{S}^{\mathrm{Gauss}}_j(S)
  =
  n\log\!\left(\operatorname{RSS}_{j,S}/n\right)
  +
  k_S\log n,
\]
up to the additive Gaussian constant independent of \(S\).  Here
\(\operatorname{RSS}_{j,S}\) is the residual sum of squares of the fitted spline
regression and \(k_S\) is the number of active spline-design columns for the
selected parents.  \ourmethodHist uses the same spline basis, search settings,
and \(k_S\log n\) penalty but replaces the Gaussian residual term by a
histogram residual likelihood.  For residuals \(\widehat N_i\), the histogram
range is split into \(B\) equal-width bins using the Freedman--Diaconis rule,
clipped to \(4\leq B\leq 64\).  If \(b(i)\) is the bin of residual \(i\),
\(c_{b(i)}\) is the corresponding bin count, \(w\) is the bin width, and
\(\alpha=0.5\), the leave-one-out density estimate is
\[
  \widehat p_{-i}(\widehat N_i)
  =
  \frac{c_{b(i)}-1+\alpha}{n-1+\alpha B}\,\frac{1}{w}.
\]
The histogram variant therefore uses
\[
  \mathcal{S}^{\mathrm{Hist}}_j(S)
  =
  2\sum_{i=1}^n -\log \widehat p_{-i}(\widehat N_i)
  +
  k_S\log n.
\]

\paragraph{Baselines.}
For \CAM, \SCORE, \NOGAM, and \DAS we used the dodiscover implementations with
their default spline settings (\(10\) splines, degree \(3\)) and the default
significance level \(0.05\).  \TOPIC was run with published implementation
and its default continuous-data GAM score.  \GLOBE was run through its published default implementation with maximum interaction order \(3\).  \NOTEARS uses
the nonlinear MLP variant with hidden layers \((10,1)\),
\(\lambda_1=\lambda_2=0.01\), \(100\) outer iterations, and edge threshold
\(0.3\).  \DAGMA uses the nonlinear MLP variant with hidden width \(10\),
\(\lambda_1=0.02\), \(\lambda_2=0.005\), L2 loss, and edge threshold \(0.3\).

\DIFFAN was run with masking enabled, batch size \(256\), ordering batch size
\(128\), learning rate \(10^{-3}\), and \(3000\) epochs.  We used the residue-free \DIFFAN variant for the scalability
experiments and the standard \DIFFAN variant for the qualitative quality
experiment.  \SCINO uses the SciNO model with batch size \(64\), ordering batch
size \(32\), evaluation batch size \(8\), \(3\) votes, learning rate \(10^{-3}\),
\(2000\) epochs, Fourier depth \(10\), batch normalization, and CAM pruning at
cutoff \(10^{-3}\).  \SDCD uses the observational-only setup with parameterized
variance, no Gumbel relaxation, stage-1 and stage-2 training for \(2000\) epochs
each, batch size \(256\), one hidden layer of width \(10\), \(15\)
power-iteration steps, validation fraction \(0.2\), and threshold \(0.1\).

\AVICI uses the pretrained \texttt{scm-v0} checkpoint and thresholds returned
edge probabilities at \(0.5\).  \CAUSCALE uses the
\texttt{OpenCausaLab/causcale-model} checkpoint
\texttt{synthetic/auprc=0.905\_migrated.ckpt}, evaluates one graph per batch,
uses at most \(500\) samples for feature subsampling, thresholds edge
probabilities at \(0.5\), and removes cycles by deleting the lowest-probability
edge on a detected cycle.  \CAPS uses \(\eta_G=\eta_H=10^{-3}\),
\(\lambda_1=\lambda_2=50\), pre-pruning, edge addition, median-bandwidth score
estimation using at most \(2048\) samples, and Stein batches of size \(64\).
\PC and \FCI use the causal-learn implementations with the kernel conditional
independence test.

%% file: tables/generated/appendix_experiment_tables.tex
% Auto-generated by paper/scripts/make_appendix_experiment_tables.py.
% Include this file from the appendix to insert all generated experiment tables.

\input{tables/generated/controlled_scalability_feature_size_n_features_25.tex}
\input{tables/generated/controlled_scalability_feature_size_n_features_50.tex}
\input{tables/generated/controlled_scalability_feature_size_n_features_100.tex}
\input{tables/generated/controlled_scalability_feature_size_n_features_200.tex}
\input{tables/generated/controlled_scalability_feature_size_n_features_400.tex}
\input{tables/generated/controlled_scalability_feature_size_n_features_800.tex}
\input{tables/generated/controlled_scalability_feature_size_n_features_1600.tex}
\input{tables/generated/controlled_scalability_sample_size_n_samples_2500.tex}
\input{tables/generated/controlled_scalability_sample_size_n_samples_5000.tex}
\input{tables/generated/controlled_scalability_sample_size_n_samples_10000.tex}
\input{tables/generated/controlled_scalability_sample_size_n_samples_20000.tex}
\input{tables/generated/controlled_scalability_sample_size_n_samples_40000.tex}
\input{tables/generated/controlled_scalability_sample_size_n_samples_80000.tex}
\input{tables/generated/controlled_scalability_sample_size_n_samples_160000.tex}
\input{tables/generated/heterogeneous_scalability_feature_size.tex}
\input{tables/generated/heterogeneous_scalability_sample_size.tex}
\input{tables/generated/qualitative_comparison_n1000_d20_deg2p0_additive.tex}
\input{tables/generated/qualitative_comparison_n1000_d20_deg2p0_neveradditive.tex}
\input{tables/generated/qualitative_comparison_n1000_d20_deg3p0_additive.tex}
\input{tables/generated/qualitative_comparison_n1000_d20_deg3p0_neveradditive.tex}
\input{tables/generated/qualitative_comparison_n1000_d20_deg4p0_additive.tex}
\input{tables/generated/qualitative_comparison_n1000_d20_deg4p0_neveradditive.tex}
\input{tables/generated/qualitative_comparison_n1000_d50_deg2p0_additive.tex}
\input{tables/generated/qualitative_comparison_n1000_d50_deg2p0_neveradditive.tex}
\input{tables/generated/qualitative_comparison_n1000_d50_deg3p0_additive.tex}
\input{tables/generated/qualitative_comparison_n1000_d50_deg3p0_neveradditive.tex}
\input{tables/generated/qualitative_comparison_n1000_d50_deg4p0_additive.tex}
\input{tables/generated/qualitative_comparison_n1000_d50_deg4p0_neveradditive.tex}
\input{tables/generated/qualitative_comparison_n2000_d20_deg2p0_additive.tex}
\input{tables/generated/qualitative_comparison_n2000_d20_deg2p0_neveradditive.tex}
\input{tables/generated/qualitative_comparison_n2000_d20_deg3p0_additive.tex}
\input{tables/generated/qualitative_comparison_n2000_d20_deg3p0_neveradditive.tex}
\input{tables/generated/qualitative_comparison_n2000_d20_deg4p0_additive.tex}
\input{tables/generated/qualitative_comparison_n2000_d20_deg4p0_neveradditive.tex}
\input{tables/generated/qualitative_comparison_n2000_d50_deg2p0_additive.tex}
\input{tables/generated/qualitative_comparison_n2000_d50_deg2p0_neveradditive.tex}
\input{tables/generated/qualitative_comparison_n2000_d50_deg3p0_additive.tex}
\input{tables/generated/qualitative_comparison_n2000_d50_deg3p0_neveradditive.tex}
\input{tables/generated/qualitative_comparison_n2000_d50_deg4p0_additive.tex}
\input{tables/generated/qualitative_comparison_n2000_d50_deg4p0_neveradditive.tex}
\input{tables/generated/real_world_benchmark.tex}
\input{tables/generated/sachs_observational_transform_benchmark.tex}